\documentclass[10pt, conference, compsocconf]{IEEEtran}
\ifCLASSINFOpdf
\else
\fi
\usepackage{graphicx}
\usepackage{bm}
\usepackage{balance}
\usepackage{amssymb}

\newcommand{\udeed}[0]{\textsc{Udeed}}
\newcommand{\lc}[0]{\textsc{Lc}}
\newcommand{\lcd}[0]{\textsc{Lcd}}
\newcommand{\lcud}[0]{\textsc{LcUd}}

\newcommand{\assemble}[0]{\textsc{Assemble}}
\newcommand{\semiboost}[0]{\textsc{SemiBoost}}
\newcommand{\ada}[0]{\textsc{AdaBoost}}

\begin{document}
%
\title{Exploiting Unlabeled Data to Enhance Ensemble Diversity}


\author{\IEEEauthorblockN{Min-Ling Zhang$^{*,\dag}$}
\IEEEauthorblockA{$^*$School of Computer Science and Engineering,\\
Southeast University, Nanjing 210096, China\\
 Email: zhangml@seu.edu.cn}
 \and \IEEEauthorblockN{Zhi-Hua Zhou$^\dag$}
\IEEEauthorblockA{$^\dag$National Key Laboratory for Novel Software Technology,\\
Nanjing University, Nanjing 210093, China\\
Email: zhouzh@lamda.nju.edu.cn} }


%


\maketitle

\begin{abstract}
Ensemble learning aims to improve generalization ability by using multiple base learners. It is
well-known that to construct a good ensemble, the base learners should be \emph{accurate} as well
as \emph{diverse}. In this paper, unlabeled data is exploited to facilitate ensemble learning by
helping augment the diversity among the base learners. Specifically, a semi-supervised ensemble
method named {\udeed} is proposed. Unlike existing semi-supervised ensemble methods where
error-prone \emph{pseudo-labels} are estimated for unlabeled data to enlarge the labeled data to
improve accuracy, {\udeed} works by maximizing accuracies of base learners on labeled data while
maximizing diversity among them on unlabeled data. Experiments show that {\udeed} can effectively
utilize unlabeled data for ensemble learning and is highly competitive to well-established
semi-supervised ensemble methods.
\end{abstract}


%
\IEEEpeerreviewmaketitle

\section{Introduction}\label{Intro}
In \textit{ensemble learning} \cite{Dietterich2000}, a number of base learners are trained and then
combined for prediction to achieve strong generalization ability. Numerous effective ensemble
methods have been proposed, such as \textsc{Boosting} \cite{FS95}, \textsc{Bagging} \cite{BRE96},
\textsc{Stacking} \cite{WOL92}, etc., and most of these methods work under the supervised setting
where the labels of training examples are known. In many real-world tasks, however, unlabeled
training examples are readily available while obtaining their labels would be fairly expensive.
\emph{Semi-supervised learning} \cite{CSZ06} is a major paradigm to exploit unlabeled data together
with labeled training data to improve learning performance automatically, without human
intervention.

This paper deals with semi-supervised ensembles, that is, ensemble learning with labeled and
unlabeled data. In contrast to the huge volume of literatures on ensemble learning and on
semi-supervised learning, only a few work has been devoted to the study of semi-supervised
ensembles. As indicated by Zhou \cite{Zhou2009a}, this was caused by the different philosophies of
the ensemble learning community and the semi-supervised learning community. The ensemble learning
community believes that it is able to boost the performance of weak learners to strong learners by
using multiple learners, and so there is no need to use unlabeled data; while the semi-supervised
learning community believes that it is able to boost the performance of weak learners to strong
learners by exploiting unlabeled data, and so there is no need to use multiple learners. However,
as Zhou indicated \cite{Zhou2009a}, there are several important reasons why ensemble learning and
semi-supervised learning are actually mutually beneficial, among which an important one is that by
considering unlabeled data it is possible to help augment the \textit{diversity} among the base
learners, as explained in the following paragraph.

It is well-known that the generalization error of an ensemble is related to the average
generalization error of the base learners and the diversity among the base learners. Generally, the
lower the average generalization error (or, the higher the average accuracy) of the base learners
and the higher the diversity among the base learners, the better the ensemble \cite{KV95}. Previous
ensemble methods work under supervised setting, trying to achieve a high average accuracy and a
high diversity by using the labeled training set. It is noteworthy, however, pursuing a high
accuracy and a high diversity may suffer from a dilemma. For example, for two classifiers which
have perfect performance on the labeled training set, they would not have diversity since there is
no difference between their predictions on the training examples. Thus, to increase the diversity
needs to sacrifice the accuracy of one classifier. However, when we have unlabeled data, we might
find that these two classifiers actually make different predictions on unlabeled data. This would
be important for ensemble design. For example, given two pairs of classifiers, $(A, B)$ and $(C,
D)$, if we know that all of them are with 100$\%$ accuracy on labeled training data, then there
will be no difference taking either the ensemble consisting of $(A, B)$ or the ensemble consisting
of $(C, D)$; however, if we find that $A$ and $B$ make the same predictions on unlabeled data,
while $C$ and $D$ make different predictions on some unlabeled data, then we will know that the
ensemble consisting of $(C, D)$ should be better. So, in contrast to previous ensemble methods
which focus on achieving both high accuracy and high diversity using only the labeled data, the use
of unlabeled data would open a promising direction for designing new ensemble methods.

In this paper, we propose the {\udeed} (\emph{Unlabeled Data to Enhance Ensemble Diversity})
approach. Experiments show that by using unlabeled data for diversity augmentation, {\udeed}
achieves much better performance than its counterpart which does not consider the usefulness of
unlabeled data. Moreover, {\udeed} also achieves highly comparable performance to other
state-of-the-art semi-supervised ensemble methods.

The rest of this paper is organized as follows. Section \ref{review} briefly reviews related work
on semi-supervised ensembles. Section \ref{approach} presents {\udeed}. Section \ref{experiments}
reports our experimental results. Finally, Section \ref{conclusion} concludes.

\section{Related Work}\label{review}
As mentioned before, in contrast to the huge volume of literatures on ensemble learning and on
semi-supervised learning, only a few work has been devoted to the study of semi-supervised
ensembles.

Zhou and Li \cite{ZL05} proposed the \textsc{Tri-training} approach which uses three classifiers
and in each round if two classifiers agree on an unlabeled instance while the third classifier
disagrees, then the two classifiers, under a certain condition, will label this unlabeled instance
for the third classifier; the three classifiers are voted to make prediction. This is a
\emph{disagreement-based} semi-supervised learning approach \cite{ZL09}, which can be viewed as a
variant of the famous \textit{co-training} method \cite{BM98}. Later, Li and Zhou \cite{LZ07}
extended \textsc{Tri-training} to \textsc{Co-forest}, by including more base classifiers and in
each round the \textit{majority teach minority} strategy is still adopted.

In addition to \textsc{Tri-training} and \textsc{Co-forest}, there are several
\emph{semi-supervised boosting} methods \cite{BDM02,CW08,AGA02,MJJL09,VJJ08}. D'Alch{\'e} Buc et
al. \cite{AGA02} proposed \textsc{SSMBoost} to handle unlabeled data within the margin cost
functional optimization framework for boosting \cite{MBBF00}, where the margin of an ensemble $H$
on unlabeled data ${\bm x}$ is defined as either $H({\bm x})^2$ or $|H({\bm x})|$. Furthermore,
\textsc{SSMBoost} requires the base learners to be semi-supervised algorithms themselves. Later,
Bennett et al. \cite{BDM02} developed {\assemble}, which labels unlabeled data ${\bm x}$ by the
current ensemble as $y={\rm sign}\left[H({\bm x})\right]$, and then iteratively puts the newly
labeled examples into the original labeled set to train a new base classifier which is then added
to $H$. Following the same margin cost functional optimization framework, Chen and Wang \cite{CW08}
added a local smoothness regularizer to the objective function used by {\assemble} to help induce
new base classifier with a more reliable self-labeling process. Other than the margin cost
functional formalization, \textsc{Mcssb} \cite{VJJ08} and {\semiboost} \cite{MJJL09} estimate the
labels of unlabeled instances by optimizing an objective function containing two terms. The first
term encodes the \textit{manifold assumption} that unlabeled instances with high similarities in
input space should share similar labels, while the other term encodes the \textit{clustering
assumption} that unlabeled instances with high similarities to a labeled example should share its
given label. The difference lies in that \textsc{Mcssb} \cite{VJJ08} implemented the objective
terms based on Bregman divergence while {\semiboost} \cite{MJJL09} implemented them with
traditional exponential loss.

A commonness of these existing semi-supervised ensemble methods is that they construct ensembles
iteratively, and in particular, the unlabeled data are exploited through assigning
\emph{pseudo-labels} for them to enlarge labeled training set. Specifically, pseudo-labels of
unlabeled instances are estimated based on the ensemble trained so far
\cite{BDM02,AGA02,LZ07,ZL05}, or with specific form of smoothness or manifold regularization
\cite{CW08,MJJL09,VJJ08}. After that, by regarding the estimated labels as their
\emph{ground-truth} labels, unlabeled instances are used in conjunction with labeled examples to
update the current ensemble iteratively.

Although various strategies have been employed to make the pseudo-labeling process more reliable,
such as by incorporating data editing \cite{Li:Zhou2005}, the estimated pseudo-labels may still be
prone to error, especially in initial training iterations where the ensemble is only moderately
accurate. In the next section we will present the {\udeed} approach. Rather than working with
pseudo-labels to enlarge labeled training set, {\udeed} utilizes unlabeled data in a different way,
i.e., help augment the \emph{diversity} among base learners.

\section{The U{\small DEED} Approach}\label{approach}

\subsection{General Formulation}
Let $\mathcal{X}=\mathcal{R}^d$ be the $d$-dimensional input space and $\mathcal{Y}=\{-1,+1\}$ be
the output space. Suppose $\mathcal{L}=\{({\bm x}_i,y_i)|$ $1\leq i\leq L\}$ contains $L$
\emph{labeled} training examples and $\mathcal{U}=\{{\bm x}_i|L+1\leq i\leq L+U\}$ contains $U$
\emph{unlabeled} training examples, where ${\bm x}_i\in\mathcal{X}$ and $y_i\in\mathcal{Y}$. In
addition, we use $\tilde{\mathcal{L}}=\{{\bm x}_i|1\leq i\leq L\}$ to denote the unlabeled data set
derived from $\mathcal{L}$.

We assume that the classifier ensemble is composed of $m$ base classifiers $\{f_k|1\leq k\leq m\}$,
where each of them takes the form $f_k:\mathcal{X}\rightarrow[-1,+1]$. Here, the real value of
$f_k({\bm x})$ corresponds to the confidence of ${\bm x}$ being positive. Accordingly, $(f_k({\bm
x})+1)/2$ can be regarded as the \emph{posteriori probability} of being positive given ${\bm x}$,
i.e. $P(y=+1|{\bm x})$.

The basic idea of {\udeed} is to maximize the fit of the classifiers on the labeled data, while
maximizing the diversity of the classifiers on the unlabeled data. Therefore, {\udeed} generates
the classifier ensemble ${\bm f}=(f_1,f_2,\cdots,f_m)$ by minimizing the following loss function:
\begin{equation}\label{loss}
  V({\bm f},\mathcal{L},\mathcal{D})=V_{emp}({\bm f},\mathcal{L})+\gamma\cdot V_{div}({\bm f},\mathcal{D})
\end{equation}
Here, the first term $V_{emp}({\bm f},\mathcal{L})$  corresponds to the \emph{empirical loss} of
${\bm f}$ on the labeled data set $\mathcal{L}$; the second term $V_{div}({\bm f},\mathcal{D})$
corresponds to the \emph{diversity loss} of ${\bm f}$ on a specified data set $\mathcal{D}$ (e.g.
$\mathcal{D}=\mathcal{U}$). Furthermore, $\gamma$ is the cost parameter balancing the importance of
the two terms.

In this paper, {\udeed} calculates the first term $V_{emp}({\bm f},\mathcal{L})$ in Eq.(\ref{loss})
as:
\begin{equation}\label{loss_acc}
  V_{emp}({\bm f},\mathcal{L})=\frac{1}{m}\cdot\sum_{k=1}^m l(f_k,\mathcal{L})
\end{equation}
Here, $l(f_k,\mathcal{L})$ measures the empirical loss of the $k$-th base classifier $f_k$ on the
labeled data set $\mathcal{L}$.

As shown in Eq.(\ref{loss}), the second term $V_{div}({\bm f},\mathcal{D})$  is used to
characterize the diversity among the based learners. However, it is well-known that diversity
measurement is not a straightforward task since there is no generally accepted formal definition
\cite{KW03}. In this paper, {\udeed} chooses to calculate $V_{div}({\bm f},\mathcal{D})$ in a novel
way as follows:
\begin{eqnarray}
  \nonumber \hspace{20pt} V_{div}({\bm f},\mathcal{D})=\frac{2}{m(m-1)}\cdot\sum_{p=1}^{m-1}\sum_{q=p+1}^m d(f_p,f_q,\mathcal{D})
\end{eqnarray}
\begin{equation}\label{loss_div}
  {\rm where}\ \ d(f_p,f_q,\mathcal{D})= \frac{1}{|\mathcal{D}|}\sum\limits_{{\bm x}\in \mathcal{D}}f_p({\bm x})f_q({\bm x})
\end{equation}
Here, $|\mathcal{D}|$ returns the cardinality of data set $\mathcal{D}$. Intuitively,
$d(f_p,f_q,\mathcal{D})$ represents the \emph{prediction difference} between any pair of base
classifiers on a specified data set $\mathcal{D}$.\footnote{As reviewed in \cite{KW03}, most
existing diversity measures are calculated based on the \emph{oracle} (correct/incorrect) outputs
of base learners, i.e. the \emph{ground-truth} labels of the data set are assumed to be known.
However, considering that examples contained in the specified data set $\mathcal{D}$ may be
\emph{unlabeled}, it is then infeasible to calculate $d(f_p,f_q,\mathcal{D})$ by directly utilizing
existing diversity measures.} In addition, the prediction difference is calculated based on the
concrete output $f({\bm x})$ instead of the signed output ${\rm sign}[f({\bm x})]$. In this way,
the \emph{prediction confidence} of each classifier other than the simple \emph{binary prediction}
is fully utilized.

Then, {\udeed} aims to find the target model ${\bm f}^{*}$ which minimizes the loss function in
Eq.(\ref{loss}):
\begin{equation}\label{optimal}
  {\bm f}^{*}=\arg\min_{{\bm f}}V({\bm f},\mathcal{L},\mathcal{D})
\end{equation}

\begin{table*}[t]
  \centering
  \renewcommand{\arraystretch}{1.4}
  \tabcolsep 0.04in
  \caption{Characteristics of the data sets (\emph{d}: dimensionality, \emph{pos.}: \#positive examples, \emph{neg.}: \#negative
  examples).}\label{data}
  \begin{small}
  \begin{tabular}{lllll|lllll|lllll|llll|lll}
\hline
\hline
data set & \multicolumn{2}{c}{\emph{d}} & \multicolumn{2}{c||}{\emph{pos.}/\emph{neg.}} & data set & \multicolumn{2}{c}{\emph{d}} & \multicolumn{2}{c||}{\emph{pos.}/\emph{neg.}} & data set & \multicolumn{2}{c}{\emph{d}} & \multicolumn{2}{c||}{\emph{pos.}/\emph{neg.}} & data set & \multicolumn{1}{c}{\emph{d}} & \multicolumn{2}{c||}{\emph{pos.}/\emph{neg.}} & data set & \multicolumn{1}{c}{\emph{d}} & \multicolumn{1}{c}{\emph{pos.}/\emph{neg.}} \\
\hline
diabetes & \multicolumn{2}{c}{8} & \multicolumn{2}{c||}{268/500} & vote & \multicolumn{2}{c}{16} & \multicolumn{2}{c||}{168/267} & ionosphere & \multicolumn{2}{c}{34} & \multicolumn{2}{c||}{255/126} & credit\_g & \multicolumn{1}{c}{61} & \multicolumn{2}{c||}{300/700} & adult & \multicolumn{1}{c}{123} & \multicolumn{1}{c}{7841/24720} \\
heart & \multicolumn{2}{c}{9} & \multicolumn{2}{c||}{120/150} & vehicle & \multicolumn{2}{c}{16} & \multicolumn{2}{c||}{218/217} & kr\_vs\_kp & \multicolumn{2}{c}{40} & \multicolumn{2}{c||}{1527/1669} & BCI & \multicolumn{1}{c}{117} & \multicolumn{2}{c||}{200/200} & web & \multicolumn{1}{c}{300} & \multicolumn{1}{c}{1479/48270} \\
wdbc & \multicolumn{2}{c}{14} & \multicolumn{2}{c||}{357/212} & hepatitis & \multicolumn{2}{c}{19} & \multicolumn{2}{c||}{123/32} & isolet & \multicolumn{2}{c}{51} & \multicolumn{2}{c||}{300/300} & Digit1 & \multicolumn{1}{c}{241} & \multicolumn{2}{c||}{734/766} & ijcnn1 & \multicolumn{1}{c}{22} & \multicolumn{1}{c}{13565/128126} \\
austra & \multicolumn{2}{c}{15} & \multicolumn{2}{c||}{307/383} & labor & \multicolumn{2}{c}{26} & \multicolumn{2}{c||}{37/20} & sonar & \multicolumn{2}{c}{60} & \multicolumn{2}{c||}{111/97} & COIL2 & \multicolumn{1}{c}{241} & \multicolumn{2}{c||}{750/750} & cod-rna & \multicolumn{1}{c}{8} & \multicolumn{1}{c}{110384/220768} \\
house & \multicolumn{2}{c}{16} & \multicolumn{2}{c||}{108/124} & ethn & \multicolumn{2}{c}{30} & \multicolumn{2}{c||}{1310/1320} & colic & \multicolumn{2}{c}{60} & \multicolumn{2}{c||}{136/232} & g241n & \multicolumn{1}{c}{241} & \multicolumn{2}{c||}{748/752} & forest & \multicolumn{1}{c}{54} & \multicolumn{1}{c}{283301/297711} \\
\hline
\hline
\end{tabular}\vspace{10pt}
\end{small}
\end{table*}

\subsection{Logistic Regression Implementation}\label{LR_Implementation}
In this paper, we employ \emph{logistic regression} to implement the base classifiers.
Specifically, each base classifier $f_k\ (1\leq k\leq m)$ is modeled as:
\begin{equation}\label{logreg}
  f_k({\bm x})=2\cdot g_k({\bm x})-1=2\cdot\frac{1}{1+e^{-({\bm w}_k^{\rm T}\cdot{\bm x}+b_k)}}-1
\end{equation}
Here, $g_k:\mathcal{X}\rightarrow[0,1]$ is the standard logistic regression function with weight
vector ${\bm w}_k\in\mathcal{R}^d$ and bias value $b_k\in\mathcal{R}$. Without loss of generality,
in the rest of this paper, $b_k$ is absorbed into ${\bm w}_k$ by appending the input space
$\mathcal{X}$ with an extra dimension fixed at value 1.

Correspondingly, the first term $V_{emp}({\bm f},\mathcal{L})$ in Eq.(\ref{loss}) is set to be the
negative \emph{binomial likelihood} function on the labeled data set $\mathcal{L}$, which is
commonly used to measure the empirical loss of logistic regression:
\begin{eqnarray}\label{loss1}
 \nonumber V_{emp}({\bm f},\mathcal{L})&=&\frac{1}{m}\cdot\sum_{k=1}^m l(f_k,\mathcal{L})\\
 \nonumber &=& \frac{1}{mL}\cdot\sum_{k=1}^m\sum_{i=1}^L-{\rm BLH}(f_k({\bm x}_i),y_i)
\end{eqnarray}
Here, the term ${\rm BLH}(f_k({\bm x}_i),y_i)$ calculates the binomial likelihood of ${\bm x}_i$
having label $y_i$, when $f_k$ serves as the classification model. Note that the probabilities of
$P(y=+1|{\bm x})$ and $P(y=-1|{\bm x})$ are modeled as $\frac{1+f_k({\bm x})}{2}$ and
$\frac{1-f_k({\bm x})}{2}$ respectively, ${\rm BLH}(f_k({\bm x}_i),y_i)$ then takes the following
form based on Eq.(\ref{logreg}):\vspace{5pt}
\begin{eqnarray}\label{loss2}
 \nonumber {\rm BLH}(f_k({\bm x}_i),y_i)\hspace{178pt}\\[10pt]
 \nonumber =\ln\left(\left(\frac{1+f_k({\bm x}_i)}{2}\right)^{\frac{1+y_i}{2}}\left(\frac{1-f_k({\bm x}_i)}{2}\right)^{\frac{1-y_i}{2}}\right)\hspace{45pt}
\end{eqnarray}
\begin{eqnarray}
=-\frac{1+y_i}{2}\ln\left(1+e^{-{\bm w}_k^{\rm T}\cdot{\bm x}_i}\right)-\frac{1-y_i}{2}\ln\left(1+e^{{\bm w}_k^{\rm T}\cdot{\bm x}_i}\right)\hspace{2pt}
\end{eqnarray}\vspace{-5pt}\\
Note that the first term $V_{emp}({\bm f},\mathcal{L})$ can also be evaluated in other ways, such
as $l_2$ loss: $\frac{1}{mL}\sum_{k=1}^m\sum_{i=1}^L\left(f_k({\bm x}_i)-y_i\right)^2$, hinge loss:
$\frac{1}{mL}\sum_{k=1}^m\sum_{i=1}^L 1-y_if_k({\bm x}_i)$, etc.

The target model ${\bm f}^*$ is found by employing \emph{gradient descent}-based techniques.
Accordingly, the gradients of $V({\bm f},\mathcal{L},\mathcal{D})$ with respect to the model
parameters ${\bf \Theta}=\{{\bm w}_k|1\leq k\leq m\}$ are determined as follows:\footnote{Note that
under logistic regression implementation, the loss function $V({\bm f},\mathcal{L},\mathcal{D})$ is
generally \emph{non-convex}, and the target model ${\bm f}^*$ returned by the gradient descent
process would correspond to a \emph{local} optimal solution.}\vspace{10pt}
\begin{eqnarray}\label{gradient}
  \nonumber \frac{\partial V}{\partial {\bf \Theta}}=\left[\frac{\partial V}{\partial {\bm w}_1},\cdots,\frac{\partial V}{\partial {\bm w}_k},\cdots,\frac{\partial V}{\partial {\bm w}_m}\right],\ \ \ {\rm where}\hspace{40pt}\\[+10pt]
    \nonumber \frac{\partial V}{\partial {\bm w}_k}=-\frac{1}{mL}\cdot\sum_{i=1}^L\frac{\partial\, {\rm BLH}(f_k({\bm x}_i),y_i)}{\partial {\bm w}_k}\hspace{72pt}\\[+10pt]
    \nonumber \hspace{40pt}+\frac{2\gamma}{m(m-1)}\cdot\sum_{k'=1,\ k' \neq k}^m\frac{\partial\, d(f_k,f_{k'},\mathcal{D})}{\partial {\bm w}_k},\ \ \ {\rm and}\hspace{4pt}\\[+10pt]
    \nonumber \frac{\partial\, {\rm BLH}(f_k({\bm x}_i),y_i)}{\partial {\bm w}_k}=\hspace{143pt}\\[+10pt]
    \nonumber \left(\frac{(1+y_i)(1-f_k({\bm x}_i))}{4}\ln\left(1+e^{-{\bm w}_k^{\rm T}\cdot {\bm x}_i}\right)\right.\hspace{65pt}\\[+10pt]
    \nonumber \left.-\frac{(1-y_i)(1+f_k({\bm x}_i))}{4}\ln\left(1+e^{{\bm w}_k^{\rm T}\cdot{\bm x}_i}\right)\right)\cdot {\bm x}_i,\ \ \ {\rm and}\hspace{11pt}
\end{eqnarray}\vspace{-5pt}
\begin{eqnarray}
    \frac{\partial\, d(f_k,f_{k'},\mathcal{D})}{\partial {\bm w}_k}=\frac{1}{2|\mathcal{D}|}\cdot\sum\limits_{{\bm x}\in\mathcal{D}}f_{k'}({\bm x})\cdot (1-f_k({\bm x})^2)\cdot {\bm x}
\end{eqnarray}\\
To initialize the ensemble, each classifier $f_k$ is learned from a \emph{bootstrapped sample} of
$\mathcal{L}$, namely $\mathcal{L}_k=\{({\bm x}^k_i,y^k_i)|1\leq i\leq L\}$, by conventional
maximum likelihood procedure. Specifically, the corresponding model parameter ${\bm w}_k$ is
obtained by minimizing the objective function $\frac{1}{2}||{\bm w}_k||^2+\lambda\cdot\sum_{i=1}^L
-{\rm BLH}(f_k({\bm x}^k_i),y^k_i)$. Here, $\lambda$ balances the model complexity and the binomial
likelihood of $f_k$ on $\mathcal{L}_k$. In this paper, $\lambda$ is set to the default value of 1.
Note that the ensemble can also be initialized in other ways, such as instantiating each ${\bm
w}_k$ with random values, etc.

As shown in Eq.(\ref{loss}), the second term $V_{div}({\bm f},\mathcal{D})$ regarding ensemble
diversity is defined on a specified data set $\mathcal{D}$. Given the labeled training set
$\mathcal{L}$ and the unlabeled training set $\mathcal{U}$, we consider three possibilities of
instantiating $\mathcal{D}$:\vspace{10pt}

\begin{itemize}
  \item $\mathcal{D}=\emptyset$: No data is employed to measure
  the diversity among base learners ($V_{div}({\bm f},\mathcal{D})$=0). The resulting
  implementation is called {\lc};\vspace{10pt}

  \item $\mathcal{D}=\tilde{\mathcal{L}}$: Labeled training examples are employed to  measure the
diversity among base learners, and the ensemble is  optimized by exploiting only $\mathcal{L}$.
The resulting implementation is called {\lcd};\vspace{10pt}

  \item $\mathcal{D}=\mathcal{U}$: Unlabeled training examples are employed to  measure the
diversity among base learners, and the ensemble is  optimized by exploiting both $\mathcal{L}$
and $\mathcal{U}$. The resulting implementation is called {\lcud};\vspace{10pt}
\end{itemize}

For {\lc} and {\lcd}, after the ensemble is initialized, a series of \emph{gradient descent} steps
are performed to optimize the model by minimizing the loss function $V({\bm
f},\mathcal{L},\mathcal{D})$ as defined in Eq.(\ref{loss}). For {\lcud} however, instead of
directly minimizing $V({\bm f},\mathcal{L},\mathcal{D})$ in the straightforward way of setting
$\mathcal{D}=\mathcal{U}$, the loss function is firstly minimized by a series of gradient descent
steps with $\mathcal{D}=\tilde{\mathcal{L}}$. After that, by using the learned model as the
\emph{starting point}, a series of gradient descent steps are further conducted to finely search
the model space with $\mathcal{D}=\mathcal{U}$. The purpose of this two-stage process is to
distinguish the \emph{priorities} of the contribution from labeled data and unlabeled
data.\footnote{Similar strategies have been adopted by some successful semi-supervised ensemble
methods \cite{MJJL09,VJJ08}, where objective terms involving labeled data are given much higher
weight than those involving unlabeled data.}

For any \emph{gradient descent}-based optimization process, it is terminated if either the loss
function $V({\bm f},\mathcal{L},\mathcal{D})$ or the diversity term $V_{div}({\bm f},\mathcal{D})$
does not decrease anymore. For each implementation of {\udeed}, the label of an unseen example
${\bm z}$ is predicted by the learned ensemble ${\bm f}^*=(f_1^*,f_2^*,\cdots,f_m^*)$ via
\emph{weighted voting}:\footnote{Compared to \emph{unweighted voting} where the label of ${\bm z}$
is predicted by ${\bm f}^*({\bm z})={\rm sign}\left[\sum_{k=1}^m {\rm sign} [f_k^*({\bm
z})]\right]$, the \emph{prediction confidence} of each base learner could be fully utilized by
weighted voting.} ${\bm f}^*({\bm z})={\rm sign}\left[\sum_{k=1}^mf_k^*({\bm z})\right]$.

Intuitively, if the ensemble does benefit from the diversity augmented by the unlabeled training
examples, {\lcud} should achieve superior performance than {\lc} and {\lcd}.

\begin{table*}[t]
  \centering
  \renewcommand{\arraystretch}{1.25}
  \caption{Predictive accuracy (mean$\pm$std.) under \emph{small-scale} ensemble
   size ($m=20$). $\bullet$/$\circ$ indicates whether U{\scriptsize DEED} is statistically superior/inferior to the
   compared algorithm (pairwise $t$-test at $95\%$ significance level).}\vspace{0pt}\label{small_scale}
\begin{small}
\begin{tabular}{llllllllllll}
\hline
\hline
 &  & \multicolumn{10}{c}{Algorithm} \\
\cline{3-12}
Data Set &  & \multicolumn{1}{c}{{\udeed}} & \multicolumn{2}{c}{} & \multicolumn{1}{c}{\textsc{Bagging}} & \multicolumn{1}{c}{} & \multicolumn{1}{c}{{\ada}} & \multicolumn{1}{c}{} & \multicolumn{1}{c}{{\assemble}} & \multicolumn{1}{c}{} & \multicolumn{1}{c}{{\semiboost}} \\
\hline
diabetes &  & \multicolumn{1}{c}{0.726$\pm$0.021} & \multicolumn{2}{c}{} & \multicolumn{1}{c}{\hspace{4pt}0.690$\pm$0.018$\bullet$} & \multicolumn{1}{c}{} & \multicolumn{1}{c}{0.728$\pm$0.029} & \multicolumn{1}{c}{} & \multicolumn{1}{c}{\hspace{4pt}0.700$\pm$0.031$\bullet$} & \multicolumn{1}{c}{} & \multicolumn{1}{c}{\hspace{4pt}0.695$\pm$0.019$\bullet$} \\
heart &  & \multicolumn{1}{c}{0.793$\pm$0.040} & \multicolumn{2}{c}{} & \multicolumn{1}{c}{\hspace{4pt}0.779$\pm$0.043$\bullet$} & \multicolumn{1}{c}{} & \multicolumn{1}{c}{\hspace{4pt}0.766$\pm$0.045$\bullet$} & \multicolumn{1}{c}{} & \multicolumn{1}{c}{\hspace{4pt}0.744$\pm$0.072$\bullet$} & \multicolumn{1}{c}{} & \multicolumn{1}{c}{0.789$\pm$0.035} \\
wdbc &  & \multicolumn{1}{c}{0.927$\pm$0.014} & \multicolumn{2}{c}{} & \multicolumn{1}{c}{\hspace{4pt}0.807$\pm$0.024$\bullet$} & \multicolumn{1}{c}{} & \multicolumn{1}{c}{0.934$\pm$0.025} & \multicolumn{1}{c}{} & \multicolumn{1}{c}{\hspace{4pt}0.898$\pm$0.070$\bullet$} & \multicolumn{1}{c}{} & \multicolumn{1}{c}{\hspace{4pt}0.793$\pm$0.028$\bullet$} \\
austra &  & \multicolumn{1}{c}{0.834$\pm$0.023} & \multicolumn{2}{c}{} & \multicolumn{1}{c}{\hspace{4pt}0.810$\pm$0.024$\bullet$} & \multicolumn{1}{c}{} & \multicolumn{1}{c}{\hspace{4pt}0.809$\pm$0.028$\bullet$} & \multicolumn{1}{c}{} & \multicolumn{1}{c}{\hspace{4pt}0.801$\pm$0.038$\bullet$} & \multicolumn{1}{c}{} & \multicolumn{1}{c}{\hspace{4pt}0.815$\pm$0.029$\bullet$} \\
house &  & \multicolumn{1}{c}{0.921$\pm$0.028} & \multicolumn{2}{c}{} & \multicolumn{1}{c}{0.922$\pm$0.027} & \multicolumn{1}{c}{} & \multicolumn{1}{c}{\hspace{4pt}0.849$\pm$0.156$\bullet$} & \multicolumn{1}{c}{} & \multicolumn{1}{c}{0.921$\pm$0.036} & \multicolumn{1}{c}{} & \multicolumn{1}{c}{0.924$\pm$0.029} \\
vote &  & \multicolumn{1}{c}{0.932$\pm$0.017} & \multicolumn{2}{c}{} & \multicolumn{1}{c}{\hspace{4pt}0.930$\pm$0.018$\bullet$} & \multicolumn{1}{c}{} & \multicolumn{1}{c}{0.906$\pm$0.106} & \multicolumn{1}{c}{} & \multicolumn{1}{c}{0.928$\pm$0.019} & \multicolumn{1}{c}{} & \multicolumn{1}{c}{0.932$\pm$0.017} \\
vehicle &  & \multicolumn{1}{c}{0.916$\pm$0.019} & \multicolumn{2}{c}{} & \multicolumn{1}{c}{0.914$\pm$0.021} & \multicolumn{1}{c}{} & \multicolumn{1}{c}{0.916$\pm$0.064} & \multicolumn{1}{c}{} & \multicolumn{1}{c}{0.921$\pm$0.029} & \multicolumn{1}{c}{} & \multicolumn{1}{c}{\hspace{4pt}0.886$\pm$0.026$\bullet$} \\
hepatitis &  & \multicolumn{1}{c}{0.800$\pm$0.042} & \multicolumn{2}{c}{} & \multicolumn{1}{c}{0.792$\pm$0.026} & \multicolumn{1}{c}{} & \multicolumn{1}{c}{\hspace{4pt}0.763$\pm$0.077$\bullet$} & \multicolumn{1}{c}{} & \multicolumn{1}{c}{0.788$\pm$0.041} & \multicolumn{1}{c}{} & \multicolumn{1}{c}{0.796$\pm$0.026} \\
labor &  & \multicolumn{1}{c}{0.809$\pm$0.072} & \multicolumn{2}{c}{} & \multicolumn{1}{c}{0.801$\pm$0.074} & \multicolumn{1}{c}{} & \multicolumn{1}{c}{\hspace{4pt}0.646$\pm$0.142$\bullet$} & \multicolumn{1}{c}{} & \multicolumn{1}{c}{\hspace{4pt}0.747$\pm$0.075$\bullet$} & \multicolumn{1}{c}{} & \multicolumn{1}{c}{0.810$\pm$0.071} \\
ethn &  & \multicolumn{1}{c}{0.944$\pm$0.007} & \multicolumn{2}{c}{} & \multicolumn{1}{c}{\hspace{4pt}0.942$\pm$0.008$\bullet$} & \multicolumn{1}{c}{} & \multicolumn{1}{c}{\hspace{4pt}0.934$\pm$0.013$\bullet$} & \multicolumn{1}{c}{} & \multicolumn{1}{c}{\hspace{4pt}0.939$\pm$0.010$\bullet$} & \multicolumn{1}{c}{} & \multicolumn{1}{c}{\hspace{4pt}0.929$\pm$0.009$\bullet$} \\
ionosphere &  & \multicolumn{1}{c}{0.795$\pm$0.043} & \multicolumn{2}{c}{} & \multicolumn{1}{c}{\hspace{4pt}0.721$\pm$0.023$\bullet$} & \multicolumn{1}{c}{} & \multicolumn{1}{c}{0.807$\pm$0.037} & \multicolumn{1}{c}{} & \multicolumn{1}{c}{\hspace{4pt}0.772$\pm$0.038$\bullet$} & \multicolumn{1}{c}{} & \multicolumn{1}{c}{\hspace{4pt}0.746$\pm$0.027$\bullet$} \\
kr\_vs\_kp &  & \multicolumn{1}{c}{0.940$\pm$0.008} & \multicolumn{2}{c}{} & \multicolumn{1}{c}{\hspace{4pt}0.938$\pm$0.008$\bullet$} & \multicolumn{1}{c}{} & \multicolumn{1}{c}{0.941$\pm$0.009} & \multicolumn{1}{c}{} & \multicolumn{1}{c}{0.942$\pm$0.010} & \multicolumn{1}{c}{} & \multicolumn{1}{c}{\hspace{4pt}0.936$\pm$0.008$\bullet$} \\
isolet &  & \multicolumn{1}{c}{0.989$\pm$0.007} & \multicolumn{2}{c}{} & \multicolumn{1}{c}{0.988$\pm$0.006} & \multicolumn{1}{c}{} & \multicolumn{1}{c}{\hspace{4pt}0.714$\pm$0.244$\bullet$} & \multicolumn{1}{c}{} & \multicolumn{1}{c}{\hspace{4pt}0.985$\pm$0.010$\bullet$} & \multicolumn{1}{c}{} & \multicolumn{1}{c}{0.989$\pm$0.005} \\
sonar &  & \multicolumn{1}{c}{0.690$\pm$0.069} & \multicolumn{2}{c}{} & \multicolumn{1}{c}{0.690$\pm$0.070} & \multicolumn{1}{c}{} & \multicolumn{1}{c}{0.701$\pm$0.063} & \multicolumn{1}{c}{} & \multicolumn{1}{c}{0.672$\pm$0.068} & \multicolumn{1}{c}{} & \multicolumn{1}{c}{0.692$\pm$0.067} \\
colic &  & \multicolumn{1}{c}{0.777$\pm$0.035} & \multicolumn{2}{c}{} & \multicolumn{1}{c}{\hspace{4pt}0.785$\pm$0.035$\circ$} & \multicolumn{1}{c}{} & \multicolumn{1}{c}{\hspace{4pt}0.747$\pm$0.039$\bullet$} & \multicolumn{1}{c}{} & \multicolumn{1}{c}{\hspace{4pt}0.748$\pm$0.037$\bullet$} & \multicolumn{1}{c}{} & \multicolumn{1}{c}{\hspace{4pt}0.765$\pm$0.041$\bullet$} \\
credit\_g &  & \multicolumn{1}{c}{0.690$\pm$0.024} & \multicolumn{2}{c}{} & \multicolumn{1}{c}{\hspace{4pt}0.710$\pm$0.019$\circ$} & \multicolumn{1}{c}{} & \multicolumn{1}{c}{\hspace{4pt}0.678$\pm$0.023$\bullet$} & \multicolumn{1}{c}{} & \multicolumn{1}{c}{0.686$\pm$0.025} & \multicolumn{1}{c}{} & \multicolumn{1}{c}{\hspace{4pt}0.702$\pm$0.019$\circ$} \\
BCI &  & \multicolumn{1}{c}{0.582$\pm$0.039} & \multicolumn{2}{c}{} & \multicolumn{1}{c}{\hspace{4pt}0.576$\pm$0.039$\bullet$} & \multicolumn{1}{c}{} & \multicolumn{1}{c}{\hspace{4pt}0.606$\pm$0.040$\circ$} & \multicolumn{1}{c}{} & \multicolumn{1}{c}{0.575$\pm$0.037} & \multicolumn{1}{c}{} & \multicolumn{1}{c}{\hspace{4pt}0.569$\pm$0.049$\bullet$} \\
Digit1 &  & \multicolumn{1}{c}{0.939$\pm$0.010} & \multicolumn{2}{c}{} & \multicolumn{1}{c}{0.940$\pm$0.009} & \multicolumn{1}{c}{} & \multicolumn{1}{c}{\hspace{4pt}0.928$\pm$0.012$\bullet$} & \multicolumn{1}{c}{} & \multicolumn{1}{c}{\hspace{4pt}0.927$\pm$0.012$\bullet$} & \multicolumn{1}{c}{} & \multicolumn{1}{c}{\hspace{4pt}0.941$\pm$0.009$\circ$} \\
COIL2 &  & \multicolumn{1}{c}{0.807$\pm$0.029} & \multicolumn{2}{c}{} & \multicolumn{1}{c}{0.809$\pm$0.028} & \multicolumn{1}{c}{} & \multicolumn{1}{c}{\hspace{4pt}0.862$\pm$0.017$\circ$} & \multicolumn{1}{c}{} & \multicolumn{1}{c}{\hspace{4pt}0.819$\pm$0.023$\circ$} & \multicolumn{1}{c}{} & \multicolumn{1}{c}{\hspace{4pt}0.823$\pm$0.021$\circ$} \\
g241n &  & \multicolumn{1}{c}{0.793$\pm$0.020} & \multicolumn{2}{c}{} & \multicolumn{1}{c}{0.794$\pm$0.018} & \multicolumn{1}{c}{} & \multicolumn{1}{c}{\hspace{4pt}0.760$\pm$0.021$\bullet$} & \multicolumn{1}{c}{} & \multicolumn{1}{c}{\hspace{4pt}0.751$\pm$0.020$\bullet$} & \multicolumn{1}{c}{} & \multicolumn{1}{c}{0.791$\pm$0.022} \\
adult &  & \multicolumn{1}{c}{0.835$\pm$0.003} & \multicolumn{2}{c}{} & \multicolumn{1}{c}{\hspace{4pt}0.844$\pm$0.002$\circ$} & \multicolumn{1}{c}{} & \multicolumn{1}{c}{\hspace{4pt}0.840$\pm$0.003$\circ$} & \multicolumn{1}{c}{} & \multicolumn{1}{c}{\hspace{4pt}0.843$\pm$0.002$\circ$} & \multicolumn{1}{c}{} & \multicolumn{1}{c}{N/A} \\
web &  & \multicolumn{1}{c}{0.981$\pm$0.001} & \multicolumn{2}{c}{} & \multicolumn{1}{c}{\hspace{4pt}0.980$\pm$0.001$\bullet$} & \multicolumn{1}{c}{} & \multicolumn{1}{c}{\hspace{4pt}0.980$\pm$0.001$\bullet$} & \multicolumn{1}{c}{} & \multicolumn{1}{c}{\hspace{4pt}0.981$\pm$0.001$\circ$} & \multicolumn{1}{c}{} & \multicolumn{1}{c}{N/A} \\
ijcnn1 &  & \multicolumn{1}{c}{0.914$\pm$0.001} & \multicolumn{2}{c}{} & \multicolumn{1}{c}{\hspace{4pt}0.906$\pm$0.001$\bullet$} & \multicolumn{1}{c}{} & \multicolumn{1}{c}{\hspace{4pt}0.910$\pm$0.004$\bullet$} & \multicolumn{1}{c}{} & \multicolumn{1}{c}{\hspace{4pt}0.906$\pm$0.001$\bullet$} & \multicolumn{1}{c}{} & \multicolumn{1}{c}{N/A} \\
cod-rna &  & \multicolumn{1}{c}{0.920$\pm$0.001} & \multicolumn{2}{c}{} & \multicolumn{1}{c}{\hspace{4pt}0.850$\pm$0.001$\bullet$} & \multicolumn{1}{c}{} & \multicolumn{1}{c}{\hspace{4pt}0.945$\pm$0.003$\circ$} & \multicolumn{1}{c}{} & \multicolumn{1}{c}{\hspace{4pt}0.851$\pm$0.002$\bullet$} & \multicolumn{1}{c}{} & \multicolumn{1}{c}{N/A} \\
forest &  & \multicolumn{1}{c}{0.706$\pm$0.002} & \multicolumn{2}{c}{} & \multicolumn{1}{c}{\hspace{4pt}0.703$\pm$0.002$\bullet$} & \multicolumn{1}{c}{} & \multicolumn{1}{c}{\hspace{4pt}0.736$\pm$0.006$\circ$} & \multicolumn{1}{c}{} & \multicolumn{1}{c}{\hspace{4pt}0.696$\pm$0.002$\bullet$} & \multicolumn{1}{c}{} & \multicolumn{1}{c}{N/A} \\
\hline
\hline
\multicolumn{1}{c}{\textbf{win/tie/loss}} &  & \multicolumn{1}{c}{$\diagup$} & \multicolumn{2}{c}{} & \multicolumn{1}{c}{\textbf{13/9/3}} & \multicolumn{1}{c}{} & \multicolumn{1}{c}{\textbf{13/7/5}} & \multicolumn{1}{c}{} & \multicolumn{1}{c}{\textbf{14/8/3}} & \multicolumn{1}{c}{} & \multicolumn{1}{c}{\textbf{9/8/3}} \\
\hline
\hline
\end{tabular}\vspace{5pt}
\end{small}
\end{table*}

\begin{table*}[t]
  \centering
  \renewcommand{\arraystretch}{1.25}
  \caption{Predictive accuracy (mean$\pm$std.) under \emph{medium-scale} ensemble
   size ($m=50$). $\bullet$/$\circ$ indicates whether U{\scriptsize DEED} is statistically superior/inferior to the
   compared algorithm (pairwise $t$-test at $95\%$ significance level).}\vspace{0pt}\label{medium_scale}
\begin{small}
\begin{tabular}{llllllllllll}
\hline
\hline
 &  & \multicolumn{10}{c}{Algorithm} \\
\cline{3-12}
Data Set &  & \multicolumn{1}{c}{{\udeed}} & \multicolumn{2}{c}{} & \multicolumn{1}{c}{\textsc{Bagging}} & \multicolumn{1}{c}{} & \multicolumn{1}{c}{{\ada}} & \multicolumn{1}{c}{} & \multicolumn{1}{c}{{\assemble}} & \multicolumn{1}{c}{} & \multicolumn{1}{c}{{\semiboost}} \\
\hline
diabetes &  & \multicolumn{1}{c}{0.710$\pm$0.020} & \multicolumn{2}{c}{} & \multicolumn{1}{c}{\hspace{4pt}0.691$\pm$0.019$\bullet$} & \multicolumn{1}{c}{} & \multicolumn{1}{c}{\hspace{4pt}0.731$\pm$0.026$\circ$} & \multicolumn{1}{c}{} & \multicolumn{1}{c}{\hspace{4pt}0.699$\pm$0.032$\bullet$} & \multicolumn{1}{c}{} & \multicolumn{1}{c}{\hspace{4pt}0.696$\pm$0.019$\bullet$} \\
heart &  & \multicolumn{1}{c}{0.794$\pm$0.033} & \multicolumn{2}{c}{} & \multicolumn{1}{c}{\hspace{4pt}0.782$\pm$0.032$\bullet$} & \multicolumn{1}{c}{} & \multicolumn{1}{c}{\hspace{4pt}0.766$\pm$0.037$\bullet$} & \multicolumn{1}{c}{} & \multicolumn{1}{c}{\hspace{4pt}0.736$\pm$0.078$\bullet$} & \multicolumn{1}{c}{} & \multicolumn{1}{c}{0.794$\pm$0.033} \\
wdbc &  & \multicolumn{1}{c}{0.885$\pm$0.017} & \multicolumn{2}{c}{} & \multicolumn{1}{c}{\hspace{4pt}0.806$\pm$0.022$\bullet$} & \multicolumn{1}{c}{} & \multicolumn{1}{c}{\hspace{4pt}0.925$\pm$0.065$\circ$} & \multicolumn{1}{c}{} & \multicolumn{1}{c}{\hspace{4pt}0.916$\pm$0.046$\circ$} & \multicolumn{1}{c}{} & \multicolumn{1}{c}{\hspace{4pt}0.816$\pm$0.033$\bullet$} \\
austra &  & \multicolumn{1}{c}{0.828$\pm$0.024} & \multicolumn{2}{c}{} & \multicolumn{1}{c}{\hspace{4pt}0.812$\pm$0.028$\bullet$} & \multicolumn{1}{c}{} & \multicolumn{1}{c}{\hspace{4pt}0.808$\pm$0.025$\bullet$} & \multicolumn{1}{c}{} & \multicolumn{1}{c}{\hspace{4pt}0.815$\pm$0.036$\bullet$} & \multicolumn{1}{c}{} & \multicolumn{1}{c}{\hspace{4pt}0.816$\pm$0.029$\bullet$} \\
house &  & \multicolumn{1}{c}{0.921$\pm$0.030} & \multicolumn{2}{c}{} & \multicolumn{1}{c}{0.920$\pm$0.030} & \multicolumn{1}{c}{} & \multicolumn{1}{c}{\hspace{4pt}0.793$\pm$0.195$\bullet$} & \multicolumn{1}{c}{} & \multicolumn{1}{c}{0.925$\pm$0.034} & \multicolumn{1}{c}{} & \multicolumn{1}{c}{\hspace{4pt}0.924$\pm$0.029$\circ$} \\
vote &  & \multicolumn{1}{c}{0.931$\pm$0.017} & \multicolumn{2}{c}{} & \multicolumn{1}{c}{\hspace{4pt}0.929$\pm$0.018$\bullet$} & \multicolumn{1}{c}{} & \multicolumn{1}{c}{\hspace{4pt}0.868$\pm$0.151$\bullet$} & \multicolumn{1}{c}{} & \multicolumn{1}{c}{0.927$\pm$0.019} & \multicolumn{1}{c}{} & \multicolumn{1}{c}{0.932$\pm$0.017} \\
vehicle &  & \multicolumn{1}{c}{0.914$\pm$0.022} & \multicolumn{2}{c}{} & \multicolumn{1}{c}{0.914$\pm$0.021} & \multicolumn{1}{c}{} & \multicolumn{1}{c}{0.914$\pm$0.088} & \multicolumn{1}{c}{} & \multicolumn{1}{c}{0.919$\pm$0.025} & \multicolumn{1}{c}{} & \multicolumn{1}{c}{\hspace{4pt}0.893$\pm$0.026$\bullet$} \\
hepatitis &  & \multicolumn{1}{c}{0.796$\pm$0.031} & \multicolumn{2}{c}{} & \multicolumn{1}{c}{0.792$\pm$0.022} & \multicolumn{1}{c}{} & \multicolumn{1}{c}{\hspace{4pt}0.737$\pm$0.106$\bullet$} & \multicolumn{1}{c}{} & \multicolumn{1}{c}{0.785$\pm$0.045} & \multicolumn{1}{c}{} & \multicolumn{1}{c}{0.797$\pm$0.027} \\
labor &  & \multicolumn{1}{c}{0.813$\pm$0.083} & \multicolumn{2}{c}{} & \multicolumn{1}{c}{\hspace{4pt}0.799$\pm$0.079$\bullet$} & \multicolumn{1}{c}{} & \multicolumn{1}{c}{\hspace{4pt}0.681$\pm$0.142$\bullet$} & \multicolumn{1}{c}{} & \multicolumn{1}{c}{\hspace{4pt}0.749$\pm$0.095$\bullet$} & \multicolumn{1}{c}{} & \multicolumn{1}{c}{0.804$\pm$0.083} \\
ethn &  & \multicolumn{1}{c}{0.944$\pm$0.006} & \multicolumn{2}{c}{} & \multicolumn{1}{c}{\hspace{4pt}0.942$\pm$0.007$\bullet$} & \multicolumn{1}{c}{} & \multicolumn{1}{c}{\hspace{4pt}0.937$\pm$0.013$\bullet$} & \multicolumn{1}{c}{} & \multicolumn{1}{c}{\hspace{4pt}0.939$\pm$0.011$\bullet$} & \multicolumn{1}{c}{} & \multicolumn{1}{c}{\hspace{4pt}0.931$\pm$0.009$\bullet$} \\
ionosphere &  & \multicolumn{1}{c}{0.797$\pm$0.042} & \multicolumn{2}{c}{} & \multicolumn{1}{c}{\hspace{4pt}0.722$\pm$0.022$\bullet$} & \multicolumn{1}{c}{} & \multicolumn{1}{c}{\hspace{4pt}0.814$\pm$0.035$\circ$} & \multicolumn{1}{c}{} & \multicolumn{1}{c}{\hspace{4pt}0.783$\pm$0.027$\bullet$} & \multicolumn{1}{c}{} & \multicolumn{1}{c}{\hspace{4pt}0.748$\pm$0.028$\bullet$} \\
kr\_vs\_kp &  & \multicolumn{1}{c}{0.939$\pm$0.008} & \multicolumn{2}{c}{} & \multicolumn{1}{c}{\hspace{4pt}0.938$\pm$0.008$\bullet$} & \multicolumn{1}{c}{} & \multicolumn{1}{c}{\hspace{4pt}0.943$\pm$0.011$\circ$} & \multicolumn{1}{c}{} & \multicolumn{1}{c}{\hspace{4pt}0.943$\pm$0.009$\circ$} & \multicolumn{1}{c}{} & \multicolumn{1}{c}{\hspace{4pt}0.935$\pm$0.008$\bullet$} \\
isolet &  & \multicolumn{1}{c}{0.989$\pm$0.006} & \multicolumn{2}{c}{} & \multicolumn{1}{c}{\hspace{4pt}0.988$\pm$0.007$\bullet$} & \multicolumn{1}{c}{} & \multicolumn{1}{c}{\hspace{4pt}0.672$\pm$0.232$\bullet$} & \multicolumn{1}{c}{} & \multicolumn{1}{c}{\hspace{4pt}0.986$\pm$0.008$\bullet$} & \multicolumn{1}{c}{} & \multicolumn{1}{c}{0.990$\pm$0.005} \\
sonar &  & \multicolumn{1}{c}{0.687$\pm$0.069} & \multicolumn{2}{c}{} & \multicolumn{1}{c}{0.690$\pm$0.072} & \multicolumn{1}{c}{} & \multicolumn{1}{c}{\hspace{4pt}0.714$\pm$0.059$\circ$} & \multicolumn{1}{c}{} & \multicolumn{1}{c}{0.679$\pm$0.070} & \multicolumn{1}{c}{} & \multicolumn{1}{c}{0.696$\pm$0.068} \\
colic &  & \multicolumn{1}{c}{0.783$\pm$0.033} & \multicolumn{2}{c}{} & \multicolumn{1}{c}{0.783$\pm$0.036} & \multicolumn{1}{c}{} & \multicolumn{1}{c}{\hspace{4pt}0.744$\pm$0.043$\bullet$} & \multicolumn{1}{c}{} & \multicolumn{1}{c}{\hspace{4pt}0.748$\pm$0.046$\bullet$} & \multicolumn{1}{c}{} & \multicolumn{1}{c}{\hspace{4pt}0.763$\pm$0.040$\bullet$} \\
credit\_g &  & \multicolumn{1}{c}{0.703$\pm$0.024} & \multicolumn{2}{c}{} & \multicolumn{1}{c}{\hspace{4pt}0.711$\pm$0.020$\circ$} & \multicolumn{1}{c}{} & \multicolumn{1}{c}{\hspace{4pt}0.674$\pm$0.026$\bullet$} & \multicolumn{1}{c}{} & \multicolumn{1}{c}{\hspace{4pt}0.689$\pm$0.025$\bullet$} & \multicolumn{1}{c}{} & \multicolumn{1}{c}{0.703$\pm$0.019} \\
BCI &  & \multicolumn{1}{c}{0.582$\pm$0.041} & \multicolumn{2}{c}{} & \multicolumn{1}{c}{0.577$\pm$0.041} & \multicolumn{1}{c}{} & \multicolumn{1}{c}{\hspace{4pt}0.620$\pm$0.043$\circ$} & \multicolumn{1}{c}{} & \multicolumn{1}{c}{0.583$\pm$0.051} & \multicolumn{1}{c}{} & \multicolumn{1}{c}{\hspace{4pt}0.572$\pm$0.045$\bullet$} \\
Digit1 &  & \multicolumn{1}{c}{0.941$\pm$0.010} & \multicolumn{2}{c}{} & \multicolumn{1}{c}{0.940$\pm$0.010} & \multicolumn{1}{c}{} & \multicolumn{1}{c}{\hspace{4pt}0.929$\pm$0.012$\bullet$} & \multicolumn{1}{c}{} & \multicolumn{1}{c}{\hspace{4pt}0.925$\pm$0.012$\bullet$} & \multicolumn{1}{c}{} & \multicolumn{1}{c}{0.941$\pm$0.009} \\
COIL2 &  & \multicolumn{1}{c}{0.808$\pm$0.027} & \multicolumn{2}{c}{} & \multicolumn{1}{c}{0.812$\pm$0.024} & \multicolumn{1}{c}{} & \multicolumn{1}{c}{\hspace{4pt}0.867$\pm$0.016$\circ$} & \multicolumn{1}{c}{} & \multicolumn{1}{c}{\hspace{4pt}0.821$\pm$0.022$\circ$} & \multicolumn{1}{c}{} & \multicolumn{1}{c}{\hspace{4pt}0.820$\pm$0.022$\circ$} \\
g241n &  & \multicolumn{1}{c}{0.796$\pm$0.019} & \multicolumn{2}{c}{} & \multicolumn{1}{c}{0.794$\pm$0.018} & \multicolumn{1}{c}{} & \multicolumn{1}{c}{\hspace{4pt}0.762$\pm$0.023$\bullet$} & \multicolumn{1}{c}{} & \multicolumn{1}{c}{\hspace{4pt}0.750$\pm$0.020$\bullet$} & \multicolumn{1}{c}{} & \multicolumn{1}{c}{\hspace{4pt}0.791$\pm$0.022$\bullet$} \\
adult &  & \multicolumn{1}{c}{0.842$\pm$0.002} & \multicolumn{2}{c}{} & \multicolumn{1}{c}{\hspace{4pt}0.844$\pm$0.002$\circ$} & \multicolumn{1}{c}{} & \multicolumn{1}{c}{\hspace{4pt}0.841$\pm$0.002$\bullet$} & \multicolumn{1}{c}{} & \multicolumn{1}{c}{\hspace{4pt}0.842$\pm$0.002$\circ$} & \multicolumn{1}{c}{} & \multicolumn{1}{c}{N/A} \\
web &  & \multicolumn{1}{c}{0.981$\pm$0.001} & \multicolumn{2}{c}{} & \multicolumn{1}{c}{\hspace{4pt}0.980$\pm$0.001$\bullet$} & \multicolumn{1}{c}{} & \multicolumn{1}{c}{0.980$\pm$0.001} & \multicolumn{1}{c}{} & \multicolumn{1}{c}{\hspace{4pt}0.981$\pm$0.001$\circ$} & \multicolumn{1}{c}{} & \multicolumn{1}{c}{N/A} \\
ijcnn1 &  & \multicolumn{1}{c}{0.907$\pm$0.001} & \multicolumn{2}{c}{} & \multicolumn{1}{c}{\hspace{4pt}0.906$\pm$0.001$\bullet$} & \multicolumn{1}{c}{} & \multicolumn{1}{c}{\hspace{4pt}0.906$\pm$0.001$\bullet$} & \multicolumn{1}{c}{} & \multicolumn{1}{c}{\hspace{4pt}0.910$\pm$0.004$\circ$} & \multicolumn{1}{c}{} & \multicolumn{1}{c}{N/A} \\
cod-rna &  & \multicolumn{1}{c}{0.891$\pm$0.001} & \multicolumn{2}{c}{} & \multicolumn{1}{c}{\hspace{4pt}0.851$\pm$0.001$\bullet$} & \multicolumn{1}{c}{} & \multicolumn{1}{c}{\hspace{4pt}0.945$\pm$0.003$\circ$} & \multicolumn{1}{c}{} & \multicolumn{1}{c}{\hspace{4pt}0.851$\pm$0.003$\bullet$} & \multicolumn{1}{c}{} & \multicolumn{1}{c}{N/A} \\
forest &  & \multicolumn{1}{c}{0.705$\pm$0.002} & \multicolumn{2}{c}{} & \multicolumn{1}{c}{\hspace{4pt}0.703$\pm$0.002$\bullet$} & \multicolumn{1}{c}{} & \multicolumn{1}{c}{\hspace{4pt}0.737$\pm$0.006$\circ$} & \multicolumn{1}{c}{} & \multicolumn{1}{c}{\hspace{4pt}0.698$\pm$0.003$\bullet$} & \multicolumn{1}{c}{} & \multicolumn{1}{c}{N/A} \\
\hline
\hline
\multicolumn{1}{c}{\textbf{win/tie/loss}} &  & \multicolumn{1}{c}{$\diagup$} & \multicolumn{2}{c}{} & \multicolumn{1}{c}{\textbf{14/9/2}} & \multicolumn{1}{c}{} & \multicolumn{1}{c}{\textbf{14/2/9}} & \multicolumn{1}{c}{} & \multicolumn{1}{c}{\textbf{13/6/6}} & \multicolumn{1}{c}{} & \multicolumn{1}{c}{\textbf{10/8/2}} \\
\hline
\hline
\end{tabular}\vspace{5pt}
\end{small}
\end{table*}

\begin{table*}[t]
  \centering
  \renewcommand{\arraystretch}{1.25}
  \caption{Predictive accuracy (mean$\pm$std.) under \emph{large-scale} ensemble
   size ($m=100$). $\bullet$/$\circ$ indicates whether U{\scriptsize DEED} is statistically superior/inferior to the
   compared algorithm (pairwise $t$-test at $95\%$ significance level).}\vspace{0pt}\label{large_scale}
\begin{small}
\begin{tabular}{llllllllllll}
\hline
\hline
 &  & \multicolumn{10}{c}{Algorithm} \\
\cline{3-12}
Data Set &  & \multicolumn{1}{c}{{\udeed}} & \multicolumn{2}{c}{} & \multicolumn{1}{c}{\textsc{Bagging}} & \multicolumn{1}{c}{} & \multicolumn{1}{c}{{\ada}} & \multicolumn{1}{c}{} & \multicolumn{1}{c}{{\assemble}} & \multicolumn{1}{c}{} & \multicolumn{1}{c}{{\semiboost}} \\
\hline
diabetes &  & \multicolumn{1}{c}{0.700$\pm$0.020} & \multicolumn{2}{c}{} & \multicolumn{1}{c}{\hspace{4pt}0.692$\pm$0.018$\bullet$} & \multicolumn{1}{c}{} & \multicolumn{1}{c}{\hspace{4pt}0.726$\pm$0.032$\circ$} & \multicolumn{1}{c}{} & \multicolumn{1}{c}{0.694$\pm$0.031} & \multicolumn{1}{c}{} & \multicolumn{1}{c}{\hspace{4pt}0.696$\pm$0.018$\bullet$} \\
heart &  & \multicolumn{1}{c}{0.790$\pm$0.035} & \multicolumn{2}{c}{} & \multicolumn{1}{c}{\hspace{4pt}0.781$\pm$0.035$\bullet$} & \multicolumn{1}{c}{} & \multicolumn{1}{c}{\hspace{4pt}0.757$\pm$0.041$\bullet$} & \multicolumn{1}{c}{} & \multicolumn{1}{c}{\hspace{4pt}0.751$\pm$0.066$\bullet$} & \multicolumn{1}{c}{} & \multicolumn{1}{c}{0.792$\pm$0.036} \\
wdbc &  & \multicolumn{1}{c}{0.852$\pm$0.021} & \multicolumn{2}{c}{} & \multicolumn{1}{c}{\hspace{4pt}0.805$\pm$0.019$\bullet$} & \multicolumn{1}{c}{} & \multicolumn{1}{c}{\hspace{4pt}0.930$\pm$0.064$\circ$} & \multicolumn{1}{c}{} & \multicolumn{1}{c}{\hspace{4pt}0.916$\pm$0.037$\circ$} & \multicolumn{1}{c}{} & \multicolumn{1}{c}{\hspace{4pt}0.825$\pm$0.030$\bullet$} \\
austra &  & \multicolumn{1}{c}{0.824$\pm$0.025} & \multicolumn{2}{c}{} & \multicolumn{1}{c}{\hspace{4pt}0.812$\pm$0.024$\bullet$} & \multicolumn{1}{c}{} & \multicolumn{1}{c}{\hspace{4pt}0.806$\pm$0.027$\bullet$} & \multicolumn{1}{c}{} & \multicolumn{1}{c}{\hspace{4pt}0.808$\pm$0.038$\bullet$} & \multicolumn{1}{c}{} & \multicolumn{1}{c}{\hspace{4pt}0.817$\pm$0.028$\bullet$} \\
house &  & \multicolumn{1}{c}{0.921$\pm$0.028} & \multicolumn{2}{c}{} & \multicolumn{1}{c}{0.921$\pm$0.029} & \multicolumn{1}{c}{} & \multicolumn{1}{c}{\hspace{4pt}0.831$\pm$0.180$\bullet$} & \multicolumn{1}{c}{} & \multicolumn{1}{c}{0.919$\pm$0.029} & \multicolumn{1}{c}{} & \multicolumn{1}{c}{\hspace{4pt}0.924$\pm$0.029$\circ$} \\
vote &  & \multicolumn{1}{c}{0.930$\pm$0.017} & \multicolumn{2}{c}{} & \multicolumn{1}{c}{0.930$\pm$0.018} & \multicolumn{1}{c}{} & \multicolumn{1}{c}{0.902$\pm$0.104} & \multicolumn{1}{c}{} & \multicolumn{1}{c}{0.926$\pm$0.020} & \multicolumn{1}{c}{} & \multicolumn{1}{c}{\hspace{4pt}0.932$\pm$0.017$\circ$} \\
vehicle &  & \multicolumn{1}{c}{0.913$\pm$0.022} & \multicolumn{2}{c}{} & \multicolumn{1}{c}{0.915$\pm$0.022} & \multicolumn{1}{c}{} & \multicolumn{1}{c}{\hspace{4pt}0.930$\pm$0.026$\circ$} & \multicolumn{1}{c}{} & \multicolumn{1}{c}{0.911$\pm$0.031} & \multicolumn{1}{c}{} & \multicolumn{1}{c}{\hspace{4pt}0.897$\pm$0.027$\bullet$} \\
hepatitis &  & \multicolumn{1}{c}{0.797$\pm$0.027} & \multicolumn{2}{c}{} & \multicolumn{1}{c}{\hspace{4pt}0.790$\pm$0.023$\bullet$} & \multicolumn{1}{c}{} & \multicolumn{1}{c}{\hspace{4pt}0.743$\pm$0.101$\bullet$} & \multicolumn{1}{c}{} & \multicolumn{1}{c}{\hspace{4pt}0.782$\pm$0.040$\bullet$} & \multicolumn{1}{c}{} & \multicolumn{1}{c}{0.797$\pm$0.026} \\
labor &  & \multicolumn{1}{c}{0.811$\pm$0.080} & \multicolumn{2}{c}{} & \multicolumn{1}{c}{0.808$\pm$0.080} & \multicolumn{1}{c}{} & \multicolumn{1}{c}{\hspace{4pt}0.683$\pm$0.146$\bullet$} & \multicolumn{1}{c}{} & \multicolumn{1}{c}{\hspace{4pt}0.756$\pm$0.098$\bullet$} & \multicolumn{1}{c}{} & \multicolumn{1}{c}{0.809$\pm$0.075} \\
ethn &  & \multicolumn{1}{c}{0.943$\pm$0.007} & \multicolumn{2}{c}{} & \multicolumn{1}{c}{0.942$\pm$0.007} & \multicolumn{1}{c}{} & \multicolumn{1}{c}{\hspace{4pt}0.938$\pm$0.012$\bullet$} & \multicolumn{1}{c}{} & \multicolumn{1}{c}{\hspace{4pt}0.939$\pm$0.011$\bullet$} & \multicolumn{1}{c}{} & \multicolumn{1}{c}{\hspace{4pt}0.932$\pm$0.008$\bullet$} \\
ionosphere &  & \multicolumn{1}{c}{0.780$\pm$0.032} & \multicolumn{2}{c}{} & \multicolumn{1}{c}{\hspace{4pt}0.721$\pm$0.023$\bullet$} & \multicolumn{1}{c}{} & \multicolumn{1}{c}{\hspace{4pt}0.812$\pm$0.037$\circ$} & \multicolumn{1}{c}{} & \multicolumn{1}{c}{0.779$\pm$0.042} & \multicolumn{1}{c}{} & \multicolumn{1}{c}{\hspace{4pt}0.747$\pm$0.027$\bullet$} \\
kr\_vs\_kp &  & \multicolumn{1}{c}{0.939$\pm$0.008} & \multicolumn{2}{c}{} & \multicolumn{1}{c}{\hspace{4pt}0.938$\pm$0.007$\bullet$} & \multicolumn{1}{c}{} & \multicolumn{1}{c}{\hspace{4pt}0.945$\pm$0.011$\circ$} & \multicolumn{1}{c}{} & \multicolumn{1}{c}{\hspace{4pt}0.944$\pm$0.008$\circ$} & \multicolumn{1}{c}{} & \multicolumn{1}{c}{\hspace{4pt}0.935$\pm$0.008$\bullet$} \\
isolet &  & \multicolumn{1}{c}{0.989$\pm$0.006} & \multicolumn{2}{c}{} & \multicolumn{1}{c}{\hspace{4pt}0.989$\pm$0.006$\bullet$} & \multicolumn{1}{c}{} & \multicolumn{1}{c}{\hspace{4pt}0.616$\pm$0.208$\bullet$} & \multicolumn{1}{c}{} & \multicolumn{1}{c}{\hspace{4pt}0.984$\pm$0.012$\bullet$} & \multicolumn{1}{c}{} & \multicolumn{1}{c}{0.990$\pm$0.005} \\
sonar &  & \multicolumn{1}{c}{0.690$\pm$0.071} & \multicolumn{2}{c}{} & \multicolumn{1}{c}{0.689$\pm$0.070} & \multicolumn{1}{c}{} & \multicolumn{1}{c}{\hspace{4pt}0.713$\pm$0.061$\circ$} & \multicolumn{1}{c}{} & \multicolumn{1}{c}{0.679$\pm$0.063} & \multicolumn{1}{c}{} & \multicolumn{1}{c}{0.696$\pm$0.069} \\
colic &  & \multicolumn{1}{c}{0.784$\pm$0.033} & \multicolumn{2}{c}{} & \multicolumn{1}{c}{0.786$\pm$0.033} & \multicolumn{1}{c}{} & \multicolumn{1}{c}{\hspace{4pt}0.741$\pm$0.041$\bullet$} & \multicolumn{1}{c}{} & \multicolumn{1}{c}{\hspace{4pt}0.745$\pm$0.051$\bullet$} & \multicolumn{1}{c}{} & \multicolumn{1}{c}{\hspace{4pt}0.763$\pm$0.042$\bullet$} \\
credit\_g &  & \multicolumn{1}{c}{0.706$\pm$0.021} & \multicolumn{2}{c}{} & \multicolumn{1}{c}{\hspace{4pt}0.711$\pm$0.021$\circ$} & \multicolumn{1}{c}{} & \multicolumn{1}{c}{\hspace{4pt}0.679$\pm$0.024$\bullet$} & \multicolumn{1}{c}{} & \multicolumn{1}{c}{\hspace{4pt}0.686$\pm$0.026$\bullet$} & \multicolumn{1}{c}{} & \multicolumn{1}{c}{0.703$\pm$0.019} \\
BCI &  & \multicolumn{1}{c}{0.580$\pm$0.041} & \multicolumn{2}{c}{} & \multicolumn{1}{c}{0.578$\pm$0.042} & \multicolumn{1}{c}{} & \multicolumn{1}{c}{\hspace{4pt}0.620$\pm$0.043$\circ$} & \multicolumn{1}{c}{} & \multicolumn{1}{c}{0.588$\pm$0.041} & \multicolumn{1}{c}{} & \multicolumn{1}{c}{0.572$\pm$0.046} \\
Digit1 &  & \multicolumn{1}{c}{0.940$\pm$0.009} & \multicolumn{2}{c}{} & \multicolumn{1}{c}{0.940$\pm$0.010} & \multicolumn{1}{c}{} & \multicolumn{1}{c}{\hspace{4pt}0.927$\pm$0.013$\bullet$} & \multicolumn{1}{c}{} & \multicolumn{1}{c}{\hspace{4pt}0.925$\pm$0.011$\bullet$} & \multicolumn{1}{c}{} & \multicolumn{1}{c}{0.941$\pm$0.009} \\
COIL2 &  & \multicolumn{1}{c}{0.807$\pm$0.027} & \multicolumn{2}{c}{} & \multicolumn{1}{c}{0.811$\pm$0.024} & \multicolumn{1}{c}{} & \multicolumn{1}{c}{\hspace{4pt}0.870$\pm$0.016$\circ$} & \multicolumn{1}{c}{} & \multicolumn{1}{c}{\hspace{4pt}0.819$\pm$0.027$\circ$} & \multicolumn{1}{c}{} & \multicolumn{1}{c}{\hspace{4pt}0.820$\pm$0.021$\circ$} \\
g241n &  & \multicolumn{1}{c}{0.795$\pm$0.018} & \multicolumn{2}{c}{} & \multicolumn{1}{c}{0.796$\pm$0.018} & \multicolumn{1}{c}{} & \multicolumn{1}{c}{\hspace{4pt}0.760$\pm$0.023$\bullet$} & \multicolumn{1}{c}{} & \multicolumn{1}{c}{\hspace{4pt}0.754$\pm$0.027$\bullet$} & \multicolumn{1}{c}{} & \multicolumn{1}{c}{0.792$\pm$0.022} \\
adult &  & \multicolumn{1}{c}{0.844$\pm$0.002} & \multicolumn{2}{c}{} & \multicolumn{1}{c}{\hspace{4pt}0.844$\pm$0.002$\circ$} & \multicolumn{1}{c}{} & \multicolumn{1}{c}{\hspace{4pt}0.840$\pm$0.002$\bullet$} & \multicolumn{1}{c}{} & \multicolumn{1}{c}{\hspace{4pt}0.843$\pm$0.002$\bullet$} & \multicolumn{1}{c}{} & \multicolumn{1}{c}{N/A} \\
web &  & \multicolumn{1}{c}{0.981$\pm$0.001} & \multicolumn{2}{c}{} & \multicolumn{1}{c}{\hspace{4pt}0.980$\pm$0.001$\bullet$} & \multicolumn{1}{c}{} & \multicolumn{1}{c}{0.980$\pm$0.002} & \multicolumn{1}{c}{} & \multicolumn{1}{c}{\hspace{4pt}0.981$\pm$0.001$\circ$} & \multicolumn{1}{c}{} & \multicolumn{1}{c}{N/A} \\
ijcnn1 &  & \multicolumn{1}{c}{0.906$\pm$0.001} & \multicolumn{2}{c}{} & \multicolumn{1}{c}{\hspace{4pt}0.905$\pm$0.004$\bullet$} & \multicolumn{1}{c}{} & \multicolumn{1}{c}{\hspace{4pt}0.906$\pm$0.001$\bullet$} & \multicolumn{1}{c}{} & \multicolumn{1}{c}{\hspace{4pt}0.906$\pm$0.001$\circ$} & \multicolumn{1}{c}{} & \multicolumn{1}{c}{N/A} \\
cod-rna &  & \multicolumn{1}{c}{0.873$\pm$0.001} & \multicolumn{2}{c}{} & \multicolumn{1}{c}{\hspace{4pt}0.851$\pm$0.001$\bullet$} & \multicolumn{1}{c}{} & \multicolumn{1}{c}{\hspace{4pt}0.945$\pm$0.003$\circ$} & \multicolumn{1}{c}{} & \multicolumn{1}{c}{\hspace{4pt}0.851$\pm$0.003$\bullet$} & \multicolumn{1}{c}{} & \multicolumn{1}{c}{N/A} \\
forest &  & \multicolumn{1}{c}{0.705$\pm$0.002} & \multicolumn{2}{c}{} & \multicolumn{1}{c}{\hspace{4pt}0.703$\pm$0.002$\bullet$} & \multicolumn{1}{c}{} & \multicolumn{1}{c}{\hspace{4pt}0.737$\pm$0.006$\circ$} & \multicolumn{1}{c}{} & \multicolumn{1}{c}{\hspace{4pt}0.698$\pm$0.003$\bullet$} & \multicolumn{1}{c}{} & \multicolumn{1}{c}{N/A} \\
\hline
\hline
\multicolumn{1}{c}{\textbf{win/tie/loss}} &  & \multicolumn{1}{c}{$\diagup$} & \multicolumn{2}{c}{} & \multicolumn{1}{c}{\textbf{12/11/2}} & \multicolumn{1}{c}{} & \multicolumn{1}{c}{\textbf{13/2/10}} & \multicolumn{1}{c}{} & \multicolumn{1}{c}{\textbf{13/7/5}} & \multicolumn{1}{c}{} & \multicolumn{1}{c}{\textbf{8/9/3}} \\
\hline
\hline
\end{tabular}\vspace{5pt}
\end{small}
\end{table*}

\section{Experiments}\label{experiments}

In this section, comparative studies between {\udeed} (i.e. {\lcud}) and other semi-supervised
ensemble methods are firstly reported. More importantly, experimental analysis on the three
different implementations of {\udeed} are further conducted to show whether unlabeled data do
benefit ensemble learning by helping augment the diversity among base learners.

Twenty-five publicly-available binary data sets are used for experiments, whose characteristics are
summarized in Table \ref{data}. Fifteen of them are from UCI Machine Learning Repository
\cite{BKM98}, five from UCI KDD Archive \cite{HB99}, four from \cite{CSZ06} and one from
\cite{LJ04}. Twenty \emph{regular-scale} data sets (left four columns) as well as five
\emph{large-scale} data sets (right column) are included. The data set size varies from 57 to
581,012, the dimensionality varies from 8 to 300, and the ratio between positive examples to
negative examples varies from 0.031 to 3.844.

For each data set, $50\%$ of them are randomly selected to form the test set $\mathcal{T}$, and the
rest is used to form the training set of $\mathcal{L}\bigcup\mathcal{U}$. The percentage of labeled
data in training set (i.e. $|\mathcal{L}|/(|\mathcal{L}|+|\mathcal{U}|)$) is set to be 0.25. For
each data set, 50 random $\mathcal{L}/\mathcal{U}/\mathcal{T}$ splits are performed. Hereafter, the
reported performance of each method corresponds to the average result out of 50 runs on different
splits.

Various ensemble sizes (i.e. $m$) are considered in the experiments: a) $m=20$ representing the
case of \emph{small-scale} ensemble; b) $m=50$ representing the case of \emph{medium-scale}
ensemble; and c) $m=100$ representing the case of \emph{large-scale} ensemble.\footnote{Preliminary
experiments show that, as the ensemble size increases from 10 to 100 within an interval of 100, the
performance of {\udeed} does not significantly change within successive ensemble sizes and tends to
converge as the ensemble size approaches 100.} In addition, as shown in Eq.(\ref{loss}), the cost
parameter $\gamma$ is set to the default value of 1. Note that better performance can be expected
if certain strategies such as cross-validation are employed to optimize the value of $\gamma$.

\begin{table*}[t]
  \centering
  \renewcommand{\arraystretch}{1.25}
  \tabcolsep 0.08in
  \caption{Accuracy improvement (mean$\pm$std.) for L{\scriptsize C}U{\scriptsize D} against L{\scriptsize C} and L{\scriptsize CD} under various ensemble
  sizes. $\bullet$/$\circ$ indicates whether L{\scriptsize C}U{\scriptsize D} is statistically superior/inferior to the compared implementation
  (pairwise $t$-test at $95\%$ significance level).}\vspace{0pt}\label{LcUd_vs_Others}
\begin{small}
\begin{tabular}{lllllllll}
\hline
\hline
 & \multicolumn{1}{c}{} & \multicolumn{7}{c}{Accuracy Improvement of {\lcud} against} \\
\cline{3-9}
 & \multicolumn{1}{c}{} & \multicolumn{3}{c}{{\lc}} & \multicolumn{1}{c}{} & \multicolumn{3}{c}{{\lcd}} \\
\cline{3-5}\cline{7-9}
Data Set & \multicolumn{1}{c}{} & \multicolumn{1}{c}{$m=20$$\hspace{5pt}$} & \multicolumn{1}{c}{$m=50$$\hspace{5pt}$} & \multicolumn{1}{c}{$m=100$$\hspace{0pt}$} & \multicolumn{1}{c}{} & \multicolumn{1}{c}{$m=20$$\hspace{5pt}$} & \multicolumn{1}{c}{$m=50$$\hspace{5pt}$} & \multicolumn{1}{c}{$m=100$$\hspace{0pt}$} \\
\hline
diabetes & \multicolumn{1}{c}{} & \multicolumn{1}{c}{0.034$\pm$0.024$\bullet$} & \multicolumn{1}{c}{0.019$\pm$0.013$\bullet$} & \multicolumn{1}{c}{0.008$\pm$0.011$\bullet$} & \multicolumn{1}{c}{} & \multicolumn{1}{c}{0.011$\pm$0.012$\bullet$} & \multicolumn{1}{c}{0.009$\pm$0.009$\bullet$} & \multicolumn{1}{c}{0.004$\pm$0.007$\bullet$} \\
heart & \multicolumn{1}{c}{} & \multicolumn{1}{c}{0.023$\pm$0.027$\bullet$} & \multicolumn{1}{c}{0.009$\pm$0.016$\bullet$} & \multicolumn{1}{c}{0.006$\pm$0.013$\bullet$} & \multicolumn{1}{c}{} & \multicolumn{1}{c}{0.009$\pm$0.016$\bullet$} & \multicolumn{1}{c}{0.003$\pm$0.010$\bullet$} & \multicolumn{1}{c}{0.004$\pm$0.009$\bullet$} \\
wdbc & \multicolumn{1}{c}{} & \multicolumn{1}{c}{0.127$\pm$0.024$\bullet$} & \multicolumn{1}{c}{0.075$\pm$0.012$\bullet$} & \multicolumn{1}{c}{0.047$\pm$0.013$\bullet$} & \multicolumn{1}{c}{} & \multicolumn{1}{c}{0.033$\pm$0.014$\bullet$} & \multicolumn{1}{c}{0.031$\pm$0.013$\bullet$} & \multicolumn{1}{c}{0.023$\pm$0.008$\bullet$} \\
austra & \multicolumn{1}{c}{} & \multicolumn{1}{c}{0.022$\pm$0.022$\bullet$} & \multicolumn{1}{c}{0.015$\pm$0.013$\bullet$} & \multicolumn{1}{c}{0.010$\pm$0.008$\bullet$} & \multicolumn{1}{c}{} & \multicolumn{1}{c}{0.004$\pm$0.012$\bullet$} & \multicolumn{1}{c}{0.006$\pm$0.008$\bullet$} & \multicolumn{1}{c}{0.005$\pm$0.005$\bullet$} \\
house & \multicolumn{1}{c}{} & \multicolumn{1}{c}{0.003$\pm$0.010$\bullet$} & \multicolumn{1}{c}{-0.001$\pm$0.005$\hspace{9pt}$} & \multicolumn{1}{c}{0.001$\pm$0.004$\bullet$} & \multicolumn{1}{c}{} & \multicolumn{1}{c}{0.002$\pm$0.007$\bullet$} & \multicolumn{1}{c}{0.000$\pm$0.004$\hspace{5pt}$} & \multicolumn{1}{c}{0.001$\pm$0.003$\bullet$} \\
vote & \multicolumn{1}{c}{} & \multicolumn{1}{c}{0.002$\pm$0.005$\bullet$} & \multicolumn{1}{c}{0.001$\pm$0.003$\bullet$} & \multicolumn{1}{c}{0.001$\pm$0.003$\bullet$} & \multicolumn{1}{c}{} & \multicolumn{1}{c}{0.001$\pm$0.004$\hspace{5pt}$} & \multicolumn{1}{c}{0.001$\pm$0.002$\bullet$} & \multicolumn{1}{c}{0.001$\pm$0.001$\bullet$} \\
vehicle & \multicolumn{1}{c}{} & \multicolumn{1}{c}{0.005$\pm$0.010$\bullet$} & \multicolumn{1}{c}{0.002$\pm$0.005$\hspace{5pt}$} & \multicolumn{1}{c}{0.001$\pm$0.004$\hspace{5pt}$} & \multicolumn{1}{c}{} & \multicolumn{1}{c}{0.003$\pm$0.007$\bullet$} & \multicolumn{1}{c}{0.001$\pm$0.005$\hspace{5pt}$} & \multicolumn{1}{c}{0.001$\pm$0.004$\hspace{5pt}$} \\
hepatitis & \multicolumn{1}{c}{} & \multicolumn{1}{c}{0.010$\pm$0.035$\hspace{5pt}$} & \multicolumn{1}{c}{0.005$\pm$0.027$\hspace{5pt}$} & \multicolumn{1}{c}{0.008$\pm$0.017$\bullet$} & \multicolumn{1}{c}{} & \multicolumn{1}{c}{0.003$\pm$0.027$\hspace{5pt}$} & \multicolumn{1}{c}{0.001$\pm$0.019$\hspace{5pt}$} & \multicolumn{1}{c}{0.005$\pm$0.012$\bullet$} \\
labor & \multicolumn{1}{c}{} & \multicolumn{1}{c}{0.003$\pm$0.071$\hspace{5pt}$} & \multicolumn{1}{c}{0.004$\pm$0.043$\hspace{5pt}$} & \multicolumn{1}{c}{0.004$\pm$0.018$\hspace{5pt}$} & \multicolumn{1}{c}{} & \multicolumn{1}{c}{-0.007$\pm$0.041$\hspace{9pt}$} & \multicolumn{1}{c}{0.007$\pm$0.032$\hspace{5pt}$} & \multicolumn{1}{c}{0.004$\pm$0.012$\bullet$} \\
ethn & \multicolumn{1}{c}{} & \multicolumn{1}{c}{0.002$\pm$0.003$\bullet$} & \multicolumn{1}{c}{0.001$\pm$0.002$\bullet$} & \multicolumn{1}{c}{0.001$\pm$0.002$\bullet$} & \multicolumn{1}{c}{} & \multicolumn{1}{c}{0.001$\pm$0.002$\bullet$} & \multicolumn{1}{c}{0.001$\pm$0.001$\bullet$} & \multicolumn{1}{c}{0.001$\pm$0.001$\bullet$} \\
ionosphere & \multicolumn{1}{c}{} & \multicolumn{1}{c}{0.073$\pm$0.049$\bullet$} & \multicolumn{1}{c}{0.076$\pm$0.049$\bullet$} & \multicolumn{1}{c}{0.057$\pm$0.035$\bullet$} & \multicolumn{1}{c}{} & \multicolumn{1}{c}{0.015$\pm$0.034$\bullet$} & \multicolumn{1}{c}{0.022$\pm$0.032$\bullet$} & \multicolumn{1}{c}{0.029$\pm$0.024$\bullet$} \\
kr\_vs\_kp & \multicolumn{1}{c}{} & \multicolumn{1}{c}{0.002$\pm$0.003$\bullet$} & \multicolumn{1}{c}{0.001$\pm$0.002$\bullet$} & \multicolumn{1}{c}{0.001$\pm$0.001$\bullet$} & \multicolumn{1}{c}{} & \multicolumn{1}{c}{0.001$\pm$0.001$\bullet$} & \multicolumn{1}{c}{0.001$\pm$0.001$\bullet$} & \multicolumn{1}{c}{0.001$\pm$0.001$\bullet$} \\
isolet & \multicolumn{1}{c}{} & \multicolumn{1}{c}{0.001$\pm$0.003$\bullet$} & \multicolumn{1}{c}{0.001$\pm$0.002$\bullet$} & \multicolumn{1}{c}{0.001$\pm$0.002$\hspace{5pt}$} & \multicolumn{1}{c}{} & \multicolumn{1}{c}{0.001$\pm$0.002$\hspace{5pt}$} & \multicolumn{1}{c}{0.001$\pm$0.001$\bullet$} & \multicolumn{1}{c}{0.001$\pm$0.001$\hspace{5pt}$} \\
sonar & \multicolumn{1}{c}{} & \multicolumn{1}{c}{0.001$\pm$0.036$\hspace{5pt}$} & \multicolumn{1}{c}{0.003$\pm$0.022$\hspace{5pt}$} & \multicolumn{1}{c}{0.001$\pm$0.015$\hspace{5pt}$} & \multicolumn{1}{c}{} & \multicolumn{1}{c}{0.002$\pm$0.016$\hspace{5pt}$} & \multicolumn{1}{c}{-0.001$\pm$0.014$\hspace{9pt}$} & \multicolumn{1}{c}{0.001$\pm$0.011$\hspace{5pt}$} \\
colic & \multicolumn{1}{c}{} & \multicolumn{1}{c}{-0.006$\pm$0.014$\circ$$\hspace{4pt}$} & \multicolumn{1}{c}{-0.003$\pm$0.012$\hspace{9pt}$} & \multicolumn{1}{c}{-0.001$\pm$0.008$\hspace{9pt}$} & \multicolumn{1}{c}{} & \multicolumn{1}{c}{-0.003$\pm$0.010$\circ$$\hspace{3pt}$} & \multicolumn{1}{c}{-0.003$\pm$0.009$\hspace{9pt}$} & \multicolumn{1}{c}{0.001$\pm$0.006$\hspace{5pt}$} \\
credit\_g & \multicolumn{1}{c}{} & \multicolumn{1}{c}{-0.019$\pm$0.017$\circ$$\hspace{4pt}$} & \multicolumn{1}{c}{-0.008$\pm$0.010$\circ$$\hspace{3pt}$} & \multicolumn{1}{c}{-0.005$\pm$0.008$\circ$$\hspace{3pt}$} & \multicolumn{1}{c}{} & \multicolumn{1}{c}{-0.009$\pm$0.010$\circ$$\hspace{3pt}$} & \multicolumn{1}{c}{-0.004$\pm$0.006$\circ$$\hspace{3pt}$} & \multicolumn{1}{c}{-0.002$\pm$0.006$\circ$$\hspace{3pt}$} \\
BCI & \multicolumn{1}{c}{} & \multicolumn{1}{c}{0.006$\pm$0.015$\bullet$} & \multicolumn{1}{c}{0.003$\pm$0.010$\hspace{5pt}$} & \multicolumn{1}{c}{0.002$\pm$0.012$\hspace{5pt}$} & \multicolumn{1}{c}{} & \multicolumn{1}{c}{0.005$\pm$0.010$\bullet$} & \multicolumn{1}{c}{0.002$\pm$0.010$\hspace{5pt}$} & \multicolumn{1}{c}{0.002$\pm$0.011$\hspace{5pt}$} \\
Digit1 & \multicolumn{1}{c}{} & \multicolumn{1}{c}{0.001$\pm$0.005$\hspace{5pt}$} & \multicolumn{1}{c}{0.001$\pm$0.002$\hspace{5pt}$} & \multicolumn{1}{c}{0.001$\pm$0.004$\hspace{5pt}$} & \multicolumn{1}{c}{} & \multicolumn{1}{c}{0.001$\pm$0.005$\hspace{5pt}$} & \multicolumn{1}{c}{0.001$\pm$0.002$\hspace{5pt}$} & \multicolumn{1}{c}{0.001$\pm$0.003$\hspace{5pt}$} \\
COIL2 & \multicolumn{1}{c}{} & \multicolumn{1}{c}{-0.001$\pm$0.016$\hspace{9pt}$} & \multicolumn{1}{c}{-0.004$\pm$0.016$\hspace{9pt}$} & \multicolumn{1}{c}{-0.003$\pm$0.015$\hspace{9pt}$} & \multicolumn{1}{c}{} & \multicolumn{1}{c}{0.001$\pm$0.005$\hspace{5pt}$} & \multicolumn{1}{c}{-0.001$\pm$0.006$\hspace{9pt}$} & \multicolumn{1}{c}{-0.002$\pm$0.007$\circ$$\hspace{3pt}$} \\
g241n & \multicolumn{1}{c}{} & \multicolumn{1}{c}{0.001$\pm$0.005$\hspace{5pt}$} & \multicolumn{1}{c}{0.001$\pm$0.004$\hspace{5pt}$} & \multicolumn{1}{c}{-0.001$\pm$0.004$\hspace{9pt}$} & \multicolumn{1}{c}{} & \multicolumn{1}{c}{-0.001$\pm$0.004$\hspace{9pt}$} & \multicolumn{1}{c}{0.001$\pm$0.004$\hspace{5pt}$} & \multicolumn{1}{c}{-0.001$\pm$0.004$\hspace{9pt}$} \\
adult & \multicolumn{1}{c}{} & \multicolumn{1}{c}{-0.009$\pm$0.002$\circ$$\hspace{3.5pt}$} & \multicolumn{1}{c}{-0.002$\pm$0.002$\circ$$\hspace{3pt}$} & \multicolumn{1}{c}{-0.001$\pm$0.001$\circ$$\hspace{3pt}$} & \multicolumn{1}{c}{} & \multicolumn{1}{c}{-0.006$\pm$0.001$\circ$$\hspace{3pt}$} & \multicolumn{1}{c}{-0.002$\pm$0.001$\circ$$\hspace{3pt}$} & \multicolumn{1}{c}{-0.001$\pm$0.001$\circ$$\hspace{3pt}$} \\
web & \multicolumn{1}{c}{} & \multicolumn{1}{c}{0.001$\pm$0.001$\bullet$} & \multicolumn{1}{c}{0.001$\pm$0.001$\bullet$} & \multicolumn{1}{c}{0.000$\pm$0.000$\hspace{5pt}$} & \multicolumn{1}{c}{} & \multicolumn{1}{c}{0.001$\pm$0.001$\bullet$} & \multicolumn{1}{c}{0.001$\pm$0.001$\bullet$} & \multicolumn{1}{c}{0.000$\pm$0.000$\hspace{5pt}$} \\
ijcnn1 & \multicolumn{1}{c}{} & \multicolumn{1}{c}{0.008$\pm$0.001$\bullet$} & \multicolumn{1}{c}{0.001$\pm$0.001$\bullet$} & \multicolumn{1}{c}{0.001$\pm$0.001$\bullet$} & \multicolumn{1}{c}{} & \multicolumn{1}{c}{0.006$\pm$0.001$\bullet$} & \multicolumn{1}{c}{0.001$\pm$0.001$\bullet$} & \multicolumn{1}{c}{0.001$\pm$0.001$\bullet$} \\
cod-rna & \multicolumn{1}{c}{} & \multicolumn{1}{c}{0.069$\pm$0.001$\bullet$} & \multicolumn{1}{c}{0.041$\pm$0.001$\bullet$} & \multicolumn{1}{c}{0.023$\pm$0.001$\bullet$} & \multicolumn{1}{c}{} & \multicolumn{1}{c}{0.022$\pm$0.001$\bullet$} & \multicolumn{1}{c}{0.018$\pm$0.001$\bullet$} & \multicolumn{1}{c}{0.011$\pm$0.001$\bullet$} \\
forest & \multicolumn{1}{c}{} & \multicolumn{1}{c}{0.003$\pm$0.001$\bullet$} & \multicolumn{1}{c}{0.002$\pm$0.001$\bullet$} & \multicolumn{1}{c}{0.001$\pm$0.001$\bullet$} & \multicolumn{1}{c}{} & \multicolumn{1}{c}{0.001$\pm$0.001$\bullet$} & \multicolumn{1}{c}{0.001$\pm$0.001$\bullet$} & \multicolumn{1}{c}{0.001$\pm$0.001$\bullet$} \\
\hline
\hline
\textbf{win/tie/loss} &  & \multicolumn{1}{c}{\textbf{16/6/3}$\hspace{3pt}$} & \multicolumn{1}{c}{\textbf{13/10/2}$\hspace{2pt}$} & \multicolumn{1}{c}{\textbf{13/10/2}$\hspace{2pt}$} & \multicolumn{1}{c}{} & \multicolumn{1}{c}{\textbf{14/8/3}$\hspace{2pt}$} & \multicolumn{1}{c}{\textbf{13/10/2}$\hspace{2pt}$} & \multicolumn{1}{c}{\textbf{14/8/3}$\hspace{2pt}$} \\
\hline
\hline
\end{tabular}\vspace{10pt}
\end{small}
\end{table*}

\subsection{Comparative Studies}
In this subsection, {\udeed} ({\lcud}) is compared with two popular ensemble methods
\textsc{Bagging} \cite{BRE96} and {\ada} \cite{FS95}, and two successful semi-supervised ensemble
methods {\assemble} \cite{BDM02} and {\semiboost} \cite{MJJL09}. For fair comparison, logistic
regression is employed as the base learner of each compared method. For {\udeed}, the maximum
number of gradient descent steps is set to 25 and the learning rate is set to 0.25. For the other
compared methods, default parameters suggested in respective literatures are adopted.

Tables \ref{small_scale} to \ref{large_scale} report the detailed experimental results under
\emph{small-scale} ($m$=20), \emph{medium-scale} ($m$=50) and \emph{large-scale} ($m$=100) ensemble
sizes respectively. {\semiboost} fails to work on the \emph{large-scale} data sets, due to its
demanding storage complexity ($\mathcal{O}(\left(|\mathcal{L}|+|\mathcal{U}|\right)^2)$) to
maintain the similarity matrix for the training examples.

On each data set, the mean predictive accuracy as well as the standard deviation of each algorithm
(out of 50 runs) are recorded. Furthermore, to statistically measure the significance of
performance difference, pairwise $t$-tests at $95\%$ significance level are conducted between the
algorithms. Specifically, whenever {\udeed} achieves significantly better/worse performance than
the compared algorithm on any data set, a win/loss is counted and a maker $\bullet$/$\circ$ is
shown. Otherwise, a tie is counted and no marker is given. The resulting win/tie/loss counts for
{\udeed} against the compared algorithms are highlighted in the last line of each table.

In summary, when the ensemble size is \emph{small} (Table \ref{small_scale}), {\udeed} is
statistically superior to \textsc{Bagging}, {\ada}, {\assemble} and {\semiboost} in $52\%$, $52\%$,
$56\%$ and $45\%$ cases, and is inferior to them in much less $12\%$, $20\%$, $12\%$ and $15\%$
cases; When the ensemble size is \emph{medium} (Table \ref{medium_scale}), {\udeed} is
statistically superior to \textsc{Bagging}, {\ada}, {\assemble} and {\semiboost} in $56\%$, $56\%$,
$52\%$ and $50\%$ cases, and is inferior to them in much less $8\%$, $36\%$, $24\%$ and $10\%$
cases; When the ensemble size is \emph{large} (Table \ref{large_scale}), {\udeed} is statistically
superior to \textsc{Bagging}, {\ada}, {\assemble} and {\semiboost} in $48\%$, $52\%$, $52\%$ and
$40\%$ cases, and is inferior to them in much less $8\%$, $40\%$, $20\%$ and $15\%$ cases. These
results indicate that {\udeed} is highly competitive to the other compared methods. Roughly
speaking, as for the time complexity, {\udeed} is slightly higher than \textsc{Bagging} and {\ada}
while fairly comparable to {\assemble} and {\semiboost}.

\subsection{The Helpfulness of Unlabeled Data}
As motivated in Section \ref{Intro}, {\udeed} aims to exploit unlabeled data to help ensemble
learning in the particular way of augmenting diversity among base learners. Therefore, in addition
to the above comparative experiments with other (semi-supervised) ensemble methods, it is more
important to show whether {\udeed} ({\lcud}) does achieve better performance than its counterparts
({\lc} and {\lcd}) which do not consider using unlabeled data for diversity augmentation.

Table \ref{LcUd_vs_Others} reports the performance improvement (i.e. increase of predictive
accuracy) of {\lcud} against {\lc} and {\lcd} under various ensemble sizes. On each data set, the
mean improved predictive accuracy as well as the standard deviation (out of 50 runs) are recorded.
In addition, to statistically measure the significance of performance difference, pairwise
$t$-tests at $95\%$ significance level are conducted. Specifically, whenever {\lcud} achieves
significantly superior/inferior performance than {\lc} or {\lcd} on any data set, a win/loss is
counted and a maker $\bullet$/$\circ$ is shown in the Table. Otherwise, a tie is counted and no
marker is given. The resulting win/tie/loss counts for {\lcud} against {\lc} and {\lcd} are
highlighted in the last line of Table \ref{LcUd_vs_Others}.

In summary, when the ensemble size is \emph{small}, {\lcud} is statistically superior to {\lc} and
{\lcd} in $64\%$ and $56\%$ cases, and is inferior to them in both only $12\%$ cases; When the
ensemble size is \emph{medium}, {\lcud} is statistically superior to {\lc} and {\lcd} in both
$52\%$ cases, and is inferior to them in both only $8\%$ cases; When the ensemble size is
\emph{large}, {\lcud} is statistically superior to {\lc} and {\lcd} in $52\%$ and $56\%$ cases, and
is inferior to them in only $8\%$ and $12\%$ cases. These results indicate that, by exploiting
unlabeled data in the specific way of helping augment ensemble diversity, {\udeed} ({\lcud) is
capable of achieving better performance than its counterparts ({\lc} and {\lcd}) which do not
consider employing unlabeled in ensemble generation.\footnote{Note that although in a number of
cases the accuracy difference between two algorithms looks rather marginal (e.g. less than $1\%$),
the difference may still be statistically significant according to the pairwise $t$-test.}

\begin{table*}[t]
  \centering
  \renewcommand{\arraystretch}{1.3}
  \caption{The win/tie/loss results for FINAL ensemble against INITIAL ensemble in terms of the
  four diversity measures under various ensemble sizes.}\label{div_wtl}\vspace{0pt}
\begin{small}
  \begin{tabular}{llllllllllllllll}
\hline
\hline
 &  & \multicolumn{14}{c}{FINAL ensemble vs. INITIAL ensemble} \\
\cline{3-16}
 &  & \multicolumn{4}{c}{$m=20$} & \multicolumn{1}{c}{} & \multicolumn{4}{c}{$m=50$} & \multicolumn{1}{c}{} & \multicolumn{4}{c}{$m=100$} \\
\cline{3-6}\cline{8-11}\cline{13-16}
Data Set &  & \multicolumn{1}{c}{DIS} & \multicolumn{1}{c}{DF} & \multicolumn{1}{c}{ENT} & \multicolumn{1}{c}{CFD} & \multicolumn{1}{c}{} & \multicolumn{1}{c}{DIS} & \multicolumn{1}{c}{DF} & \multicolumn{1}{c}{ENT} & \multicolumn{1}{c}{CFD} & \multicolumn{1}{c}{} & \multicolumn{1}{c}{DIS} & \multicolumn{1}{c}{DF} & \multicolumn{1}{c}{ENT} & \multicolumn{1}{c}{CFD} \\
\hline
diabetes &  & \multicolumn{1}{c}{win} & \multicolumn{1}{c}{win} & \multicolumn{1}{c}{win} & \multicolumn{1}{c}{win} & \multicolumn{1}{c}{} & \multicolumn{1}{c}{win} & \multicolumn{1}{c}{win} & \multicolumn{1}{c}{win} & \multicolumn{1}{c}{win} & \multicolumn{1}{c}{} & \multicolumn{1}{c}{win} & \multicolumn{1}{c}{win} & \multicolumn{1}{c}{win} & \multicolumn{1}{c}{win} \\
heart &  & \multicolumn{1}{c}{loss} & \multicolumn{1}{c}{win} & \multicolumn{1}{c}{loss} & \multicolumn{1}{c}{tie} & \multicolumn{1}{c}{} & \multicolumn{1}{c}{loss} & \multicolumn{1}{c}{win} & \multicolumn{1}{c}{loss} & \multicolumn{1}{c}{loss} & \multicolumn{1}{c}{} & \multicolumn{1}{c}{loss} & \multicolumn{1}{c}{win} & \multicolumn{1}{c}{loss} & \multicolumn{1}{c}{loss} \\
wdbc &  & \multicolumn{1}{c}{tie} & \multicolumn{1}{c}{win} & \multicolumn{1}{c}{tie} & \multicolumn{1}{c}{tie} & \multicolumn{1}{c}{} & \multicolumn{1}{c}{tie} & \multicolumn{1}{c}{tie} & \multicolumn{1}{c}{tie} & \multicolumn{1}{c}{tie} & \multicolumn{1}{c}{} & \multicolumn{1}{c}{tie} & \multicolumn{1}{c}{win} & \multicolumn{1}{c}{tie} & \multicolumn{1}{c}{tie} \\
austra &  & \multicolumn{1}{c}{loss} & \multicolumn{1}{c}{win} & \multicolumn{1}{c}{loss} & \multicolumn{1}{c}{tie} & \multicolumn{1}{c}{} & \multicolumn{1}{c}{loss} & \multicolumn{1}{c}{win} & \multicolumn{1}{c}{loss} & \multicolumn{1}{c}{tie} & \multicolumn{1}{c}{} & \multicolumn{1}{c}{loss} & \multicolumn{1}{c}{win} & \multicolumn{1}{c}{loss} & \multicolumn{1}{c}{loss} \\
house &  & \multicolumn{1}{c}{win} & \multicolumn{1}{c}{win} & \multicolumn{1}{c}{win} & \multicolumn{1}{c}{win} & \multicolumn{1}{c}{} & \multicolumn{1}{c}{win} & \multicolumn{1}{c}{win} & \multicolumn{1}{c}{win} & \multicolumn{1}{c}{win} & \multicolumn{1}{c}{} & \multicolumn{1}{c}{win} & \multicolumn{1}{c}{win} & \multicolumn{1}{c}{win} & \multicolumn{1}{c}{win} \\
vote &  & \multicolumn{1}{c}{win} & \multicolumn{1}{c}{win} & \multicolumn{1}{c}{win} & \multicolumn{1}{c}{win} & \multicolumn{1}{c}{} & \multicolumn{1}{c}{win} & \multicolumn{1}{c}{win} & \multicolumn{1}{c}{win} & \multicolumn{1}{c}{win} & \multicolumn{1}{c}{} & \multicolumn{1}{c}{win} & \multicolumn{1}{c}{win} & \multicolumn{1}{c}{win} & \multicolumn{1}{c}{win} \\
vehicle &  & \multicolumn{1}{c}{tie} & \multicolumn{1}{c}{tie} & \multicolumn{1}{c}{tie} & \multicolumn{1}{c}{tie} & \multicolumn{1}{c}{} & \multicolumn{1}{c}{loss} & \multicolumn{1}{c}{tie} & \multicolumn{1}{c}{tie} & \multicolumn{1}{c}{tie} & \multicolumn{1}{c}{} & \multicolumn{1}{c}{win} & \multicolumn{1}{c}{tie} & \multicolumn{1}{c}{win} & \multicolumn{1}{c}{tie} \\
hepatitis &  & \multicolumn{1}{c}{win} & \multicolumn{1}{c}{tie} & \multicolumn{1}{c}{win} & \multicolumn{1}{c}{win} & \multicolumn{1}{c}{} & \multicolumn{1}{c}{win} & \multicolumn{1}{c}{win} & \multicolumn{1}{c}{win} & \multicolumn{1}{c}{win} & \multicolumn{1}{c}{} & \multicolumn{1}{c}{win} & \multicolumn{1}{c}{win} & \multicolumn{1}{c}{win} & \multicolumn{1}{c}{win} \\
labor &  & \multicolumn{1}{c}{tie} & \multicolumn{1}{c}{tie} & \multicolumn{1}{c}{tie} & \multicolumn{1}{c}{tie} & \multicolumn{1}{c}{} & \multicolumn{1}{c}{win} & \multicolumn{1}{c}{win} & \multicolumn{1}{c}{win} & \multicolumn{1}{c}{tie} & \multicolumn{1}{c}{} & \multicolumn{1}{c}{win} & \multicolumn{1}{c}{win} & \multicolumn{1}{c}{win} & \multicolumn{1}{c}{tie} \\
ethn &  & \multicolumn{1}{c}{win} & \multicolumn{1}{c}{win} & \multicolumn{1}{c}{win} & \multicolumn{1}{c}{win} & \multicolumn{1}{c}{} & \multicolumn{1}{c}{loss} & \multicolumn{1}{c}{tie} & \multicolumn{1}{c}{tie} & \multicolumn{1}{c}{tie} & \multicolumn{1}{c}{} & \multicolumn{1}{c}{win} & \multicolumn{1}{c}{win} & \multicolumn{1}{c}{win} & \multicolumn{1}{c}{tie} \\
ionosphere &  & \multicolumn{1}{c}{win} & \multicolumn{1}{c}{win} & \multicolumn{1}{c}{win} & \multicolumn{1}{c}{win} & \multicolumn{1}{c}{} & \multicolumn{1}{c}{win} & \multicolumn{1}{c}{win} & \multicolumn{1}{c}{win} & \multicolumn{1}{c}{win} & \multicolumn{1}{c}{} & \multicolumn{1}{c}{win} & \multicolumn{1}{c}{win} & \multicolumn{1}{c}{win} & \multicolumn{1}{c}{win} \\
kr\_vs\_kp &  & \multicolumn{1}{c}{win} & \multicolumn{1}{c}{win} & \multicolumn{1}{c}{win} & \multicolumn{1}{c}{win} & \multicolumn{1}{c}{} & \multicolumn{1}{c}{win} & \multicolumn{1}{c}{win} & \multicolumn{1}{c}{win} & \multicolumn{1}{c}{win} & \multicolumn{1}{c}{} & \multicolumn{1}{c}{win} & \multicolumn{1}{c}{win} & \multicolumn{1}{c}{win} & \multicolumn{1}{c}{win} \\
isolet &  & \multicolumn{1}{c}{win} & \multicolumn{1}{c}{tie} & \multicolumn{1}{c}{win} & \multicolumn{1}{c}{tie} & \multicolumn{1}{c}{} & \multicolumn{1}{c}{win} & \multicolumn{1}{c}{loss} & \multicolumn{1}{c}{win} & \multicolumn{1}{c}{tie} & \multicolumn{1}{c}{} & \multicolumn{1}{c}{win} & \multicolumn{1}{c}{loss} & \multicolumn{1}{c}{win} & \multicolumn{1}{c}{tie} \\
sonar &  & \multicolumn{1}{c}{loss} & \multicolumn{1}{c}{tie} & \multicolumn{1}{c}{loss} & \multicolumn{1}{c}{loss} & \multicolumn{1}{c}{} & \multicolumn{1}{c}{loss} & \multicolumn{1}{c}{tie} & \multicolumn{1}{c}{loss} & \multicolumn{1}{c}{tie} & \multicolumn{1}{c}{} & \multicolumn{1}{c}{loss} & \multicolumn{1}{c}{tie} & \multicolumn{1}{c}{loss} & \multicolumn{1}{c}{tie} \\
colic &  & \multicolumn{1}{c}{win} & \multicolumn{1}{c}{loss} & \multicolumn{1}{c}{win} & \multicolumn{1}{c}{win} & \multicolumn{1}{c}{} & \multicolumn{1}{c}{win} & \multicolumn{1}{c}{tie} & \multicolumn{1}{c}{win} & \multicolumn{1}{c}{tie} & \multicolumn{1}{c}{} & \multicolumn{1}{c}{win} & \multicolumn{1}{c}{tie} & \multicolumn{1}{c}{win} & \multicolumn{1}{c}{tie} \\
credit\_g &  & \multicolumn{1}{c}{win} & \multicolumn{1}{c}{loss} & \multicolumn{1}{c}{win} & \multicolumn{1}{c}{win} & \multicolumn{1}{c}{} & \multicolumn{1}{c}{win} & \multicolumn{1}{c}{loss} & \multicolumn{1}{c}{win} & \multicolumn{1}{c}{win} & \multicolumn{1}{c}{} & \multicolumn{1}{c}{win} & \multicolumn{1}{c}{loss} & \multicolumn{1}{c}{win} & \multicolumn{1}{c}{win} \\
BCI &  & \multicolumn{1}{c}{win} & \multicolumn{1}{c}{win} & \multicolumn{1}{c}{win} & \multicolumn{1}{c}{win} & \multicolumn{1}{c}{} & \multicolumn{1}{c}{win} & \multicolumn{1}{c}{win} & \multicolumn{1}{c}{win} & \multicolumn{1}{c}{win} & \multicolumn{1}{c}{} & \multicolumn{1}{c}{win} & \multicolumn{1}{c}{win} & \multicolumn{1}{c}{win} & \multicolumn{1}{c}{win} \\
Digit1 &  & \multicolumn{1}{c}{win} & \multicolumn{1}{c}{win} & \multicolumn{1}{c}{win} & \multicolumn{1}{c}{win} & \multicolumn{1}{c}{} & \multicolumn{1}{c}{win} & \multicolumn{1}{c}{win} & \multicolumn{1}{c}{win} & \multicolumn{1}{c}{win} & \multicolumn{1}{c}{} & \multicolumn{1}{c}{win} & \multicolumn{1}{c}{win} & \multicolumn{1}{c}{win} & \multicolumn{1}{c}{win} \\
COIL2 &  & \multicolumn{1}{c}{win} & \multicolumn{1}{c}{win} & \multicolumn{1}{c}{win} & \multicolumn{1}{c}{win} & \multicolumn{1}{c}{} & \multicolumn{1}{c}{tie} & \multicolumn{1}{c}{win} & \multicolumn{1}{c}{tie} & \multicolumn{1}{c}{win} & \multicolumn{1}{c}{} & \multicolumn{1}{c}{tie} & \multicolumn{1}{c}{win} & \multicolumn{1}{c}{tie} & \multicolumn{1}{c}{win} \\
g241n &  & \multicolumn{1}{c}{tie} & \multicolumn{1}{c}{loss} & \multicolumn{1}{c}{tie} & \multicolumn{1}{c}{tie} & \multicolumn{1}{c}{} & \multicolumn{1}{c}{tie} & \multicolumn{1}{c}{tie} & \multicolumn{1}{c}{tie} & \multicolumn{1}{c}{tie} & \multicolumn{1}{c}{} & \multicolumn{1}{c}{tie} & \multicolumn{1}{c}{loss} & \multicolumn{1}{c}{tie} & \multicolumn{1}{c}{tie} \\
adult &  & \multicolumn{1}{c}{win} & \multicolumn{1}{c}{loss} & \multicolumn{1}{c}{win} & \multicolumn{1}{c}{win} & \multicolumn{1}{c}{} & \multicolumn{1}{c}{win} & \multicolumn{1}{c}{loss} & \multicolumn{1}{c}{win} & \multicolumn{1}{c}{win} & \multicolumn{1}{c}{} & \multicolumn{1}{c}{win} & \multicolumn{1}{c}{win} & \multicolumn{1}{c}{win} & \multicolumn{1}{c}{win} \\
web &  & \multicolumn{1}{c}{win} & \multicolumn{1}{c}{win} & \multicolumn{1}{c}{win} & \multicolumn{1}{c}{win} & \multicolumn{1}{c}{} & \multicolumn{1}{c}{win} & \multicolumn{1}{c}{win} & \multicolumn{1}{c}{win} & \multicolumn{1}{c}{win} & \multicolumn{1}{c}{} & \multicolumn{1}{c}{win} & \multicolumn{1}{c}{win} & \multicolumn{1}{c}{win} & \multicolumn{1}{c}{win} \\
ijcnn1 &  & \multicolumn{1}{c}{loss} & \multicolumn{1}{c}{loss} & \multicolumn{1}{c}{loss} & \multicolumn{1}{c}{loss} & \multicolumn{1}{c}{} & \multicolumn{1}{c}{loss} & \multicolumn{1}{c}{loss} & \multicolumn{1}{c}{loss} & \multicolumn{1}{c}{loss} & \multicolumn{1}{c}{} & \multicolumn{1}{c}{loss} & \multicolumn{1}{c}{loss} & \multicolumn{1}{c}{loss} & \multicolumn{1}{c}{loss} \\
cod-rna &  & \multicolumn{1}{c}{tie} & \multicolumn{1}{c}{win} & \multicolumn{1}{c}{tie} & \multicolumn{1}{c}{win} & \multicolumn{1}{c}{} & \multicolumn{1}{c}{tie} & \multicolumn{1}{c}{win} & \multicolumn{1}{c}{tie} & \multicolumn{1}{c}{tie} & \multicolumn{1}{c}{} & \multicolumn{1}{c}{win} & \multicolumn{1}{c}{win} & \multicolumn{1}{c}{tie} & \multicolumn{1}{c}{tie} \\
forest &  & \multicolumn{1}{c}{tie} & \multicolumn{1}{c}{tie} & \multicolumn{1}{c}{tie} & \multicolumn{1}{c}{tie} & \multicolumn{1}{c}{} & \multicolumn{1}{c}{tie} & \multicolumn{1}{c}{tie} & \multicolumn{1}{c}{tie} & \multicolumn{1}{c}{tie} & \multicolumn{1}{c}{} & \multicolumn{1}{c}{tie} & \multicolumn{1}{c}{tie} & \multicolumn{1}{c}{tie} & \multicolumn{1}{c}{tie} \\
\hline
\hline
\textbf{win/tie/loss} &  & \multicolumn{1}{c}{\textbf{\scriptsize 15/6/4}} & \multicolumn{1}{c}{\textbf{\scriptsize 14/6/5}} & \multicolumn{1}{c}{\textbf{\scriptsize 15/6/4}} & \multicolumn{1}{c}{\textbf{\scriptsize 15/8/2}} & \multicolumn{1}{c}{} & \multicolumn{1}{c}{\textbf{\scriptsize 14/5/6}} & \multicolumn{1}{c}{\textbf{\scriptsize 14/7/4}} & \multicolumn{1}{c}{\textbf{\scriptsize 14/7/4}} & \multicolumn{1}{c}{\textbf{\scriptsize 12/11/2}} & \multicolumn{1}{c}{} & \multicolumn{1}{c}{\textbf{\scriptsize 17/4/4}} & \multicolumn{1}{c}{\textbf{\scriptsize 17/4/4}} & \multicolumn{1}{c}{\textbf{\scriptsize 16/5/4}} & \multicolumn{1}{c}{\textbf{\scriptsize 12/10/3}} \\
\hline
\hline
\end{tabular}\vspace{10pt}
\end{small}
\end{table*}

\subsection{Diversity Analysis}
To clearly verify that {\udeed} ({\lcud}) does increase the diversity among base learners after
generating ensemble by utilizing unlabeled data, additional experiments are analyzed in this
subsection based on several existing diversity measures. Specifically, four diversity measures
summarized in \cite{KW03} are considered, whose values are calculated based on the \emph{oracle}
(correct/incorrect) outputs of base learners.

Suppose $m$ denotes the number of base classifiers in the ensemble and $N$ denotes the number of
examples in the test set $\mathcal{T}$. In addition, let ${\bm O}=[o_{ij}]_{m\times N}$ be the
oracle output matrix. Here, $o_{ij}=1$ if the $i$-th base learner correctly classifies the $j$-th
test example ($1\leq i\leq m,\,1\leq j\leq N$). Otherwise, $o_{ij}=0$. The formal definitions of
the
four diversity measures are as follows:\\[15pt]
$\bullet$ \emph{Disagreement measure} (DIS):\vspace{5pt}
\begin{eqnarray}
    \nonumber {\rm DIS}=\frac{2}{m(m-1)}\sum_{i=1}^{m-1}\sum_{k=i+1}^m\,{\rm dis}_{ik},\ \ {\rm where}\hspace{20pt}\\[5pt]
    \nonumber {\rm dis}_{ik}=\frac{\sum_{j=1}^N o_{ij}\cdot(1-o_{kj})+\sum_{j=1}^N (1-o_{ij})\cdot o_{kj}}{N}
\end{eqnarray}\\[5pt]
$\bullet$ \emph{Double-fault measure} (DF):\vspace{5pt}
\begin{eqnarray}
    \nonumber {\rm DF}=\frac{2}{m(m-1)}\sum_{i=1}^{m-1}\sum_{k=i+1}^m\,{\rm df}_{ik},\ \ {\rm where}\\[5pt]
    \nonumber {\rm df}_{ik}=\frac{\sum_{j=1}^N (1-o_{ij})\cdot(1-o_{kj})}{N}\hspace{15pt}
\end{eqnarray}\\[5pt]
$\bullet$ \emph{Entropy measure} (ENT):\vspace{5pt}
\begin{eqnarray}
    \nonumber {\rm ENT}=\frac{1}{N}\sum_{j=1}^N\frac{1}{m-\lceil m/2\rceil}\min\left\{\sum_{i=1}^m o_{ij},m-\sum_{i=1}^m o_{ij}\right\}
\end{eqnarray}\\[5pt]
$\bullet$ \emph{Coincident failure diversity} (CFD):\vspace{5pt}
\begin{eqnarray}
\nonumber {\rm CFD} = \left\{ {\begin{array}{*{20}c}
   {0,}\hspace{82pt} p_0=1.0\vspace{10pt}  \\
   \frac{1}{1-p_0}\sum\nolimits_{i=1}^m\frac{m-i}{m-1}p_i,\hspace{10pt} p_0<1.0\\ \end{array} } \right.,\ \ {\rm where}\\[10pt]
    \nonumber p_i=\frac{\sum_{j=1}^N\ {\bf 1}_{\left[i=\sum_{k=1}^m (1-o_{kj})\right]}}{N},\hspace{10pt}(0\leq i\leq m)
\end{eqnarray}\vspace{0pt}

Here, DIS and DF are \emph{pairwise} measures while ENT and CFD are \emph{non-pairwise} measures.
In addition, 1-DF is used instead of DF such that for all the measures, the \emph{greater} the
value the \emph{higher} the diversity. All the four measures vary between 0 and 1.

Table \ref{div_wtl} compares {\udeed}'s \emph{initial} diversity after ensemble initialization with
its \emph{final} diversity after ensemble learning under various ensemble sizes. For each data set,
pairwise $t$-tests at $95\%$ significance level are conducted between the initial and the final
ensemble diversities. Whenever the final ensemble achieves significantly higher/lower diversity
than the initial one, a win/loss is recorded. Otherwise, a tie is recorded. The resulting
win/tie/loss counts are highlighted in the last line of Table \ref{div_wtl}.

In summary, when the ensemble size is \emph{small}, {\udeed} statistically increases the initial
ensemble diversity in $60\%$ (DIS), $56\%$ (DF), $60\%$ (ENT) and $60\%$ (CFD) cases, but decreases
the initial ensemble diversity in only $16\%$ (DIS), $20\%$ (DF), $16\%$ (ENT) and $8\%$ (CFD)
cases.

When the ensemble size is \emph{medium}, {\udeed} statistically increases the initial ensemble
diversity in $56\%$ (DIS), $56\%$ (DF), $56\%$ (ENT) and $48\%$ (CFD) cases, but decreases the
initial ensemble diversity in only $24\%$ (DIS), $16\%$ (DF), $16\%$ (ENT) and $8\%$ (CFD) cases;

Finally, when the ensemble size is \emph{large}, {\udeed} statistically increases the initial
ensemble diversity in $68\%$ (DIS), $68\%$ (DF), $64\%$ (ENT) and $48\%$ (CFD) cases, but decreases
the initial ensemble diversity in only $16\%$ (DIS), $16\%$ (DF), $16\%$ (ENT) and $12\%$ (CFD)
cases.

These results clearly verify that {\udeed} can effectively exploit unlabeled data to help augment
ensemble diversity.

\section{Conclusion}\label{conclusion}

Previous ensemble methods try to obtain a high accuracy of base learners and high diversity among
base learners by considering only labeled data. There were some studies on using unlabeled data,
but focusing on using unlabeled data to improve accuracy. The major contribution of our work is to
use unlabeled data to augment diversity, which suggests a new direction for ensemble design.
Specifically, a novel semi-supervised ensemble method named {\udeed} is proposed, which works by
maximizing accuracy on labeled data while maximizing diversity on unlabeled data.

Experiments show that: a) {\udeed} achieves highly comparable performance against other successful
semi-supervised ensemble methods; b) {\udeed} does benefit from unlabeled data by using them to
augment the diversity among base learners. In the future, it is interesting to see whether {\udeed}
works well with other base learners. It would be insightful to analyze why {\udeed} can achieve
good performance theoretically. Furthermore, designing other ensemble methods by exploiting
unlabeled data to augment ensemble diversity gracefully is a direction very worth studying.


\section*{Acknowledgment}

The authors wish to thank the anonymous reviewers for their helpful comments in improving this
paper. This work was supported by the National Science Foundation of China (60635030, 60805022),
the National Fundamental Research Program of China (2010CB327903), the Ph.D. Programs Foundation of
Ministry of Education of China (200802941009), the Jiangsu Science Foundation (BK2008018) and the
Jiangsu 333 Program.

\bibliographystyle{IEEEtranS}
\bibliography{UDEED}

\begin{thebibliography}{10}
\providecommand{\url}[1]{#1}
\csname url@samestyle\endcsname
\providecommand{\newblock}{\relax}
\providecommand{\bibinfo}[2]{#2}
\providecommand{\BIBentrySTDinterwordspacing}{\spaceskip=0pt\relax}
\providecommand{\BIBentryALTinterwordstretchfactor}{4}
\providecommand{\BIBentryALTinterwordspacing}{\spaceskip=\fontdimen2\font plus
\BIBentryALTinterwordstretchfactor\fontdimen3\font minus
  \fontdimen4\font\relax}
\providecommand{\BIBforeignlanguage}[2]{{%
\expandafter\ifx\csname l@#1\endcsname\relax
\typeout{** WARNING: IEEEtranS.bst: No hyphenation pattern has been}%
\typeout{** loaded for the language `#1'. Using the pattern for}%
\typeout{** the default language instead.}%
\else
\language=\csname l@#1\endcsname
\fi
#2}}
\providecommand{\BIBdecl}{\relax}
\BIBdecl

\bibitem{BDM02}
K.~Bennett, A.~Demiriz, and R.~Maclin, ``Exploiting unlabeled data in ensemble
  methods,'' in \emph{Proceedings of the 8th ACM SIGKDD International
  Conference on Knowledge Discovery and Data Mining}, Edmonton, Canada, 2002,
  pp. 289--296.

\bibitem{BKM98}
C.~Blake, E.~Keogh, and C.~J. Merz, ``{UCI} repository of machine learning
  databases [http://www.ics.uci.edu/ {\~{}}mlearn/mlrepository.html],''
  Department of Information and Computer Science, University of California,
  Irvine, CA, Tech. Rep., 1998.

\bibitem{BM98}
A.~Blum and T.~Mitchell, ``Combining labeled and unlabeled data with
  co-training,'' in \emph{Proceedings of the 11th Annual Conference on
  Computational Learning Theory}, Madison, WI, 1998, pp. 92--100.

\bibitem{BRE96}
L.~Breiman, ``Bagging predictors,'' \emph{Machine Learning}, vol.~24, no.~2,
  pp. 123--140, 1996.

\bibitem{CSZ06}
O.~Chapelle, B.~Sch{\"o}lkopf, and A.~Zien, \emph{Semi-Supervised
  Learning}.\hskip 1em plus 0.5em minus 0.4em\relax Cambridge, MA: MIT Press,
  2006.

\bibitem{CW08}
K.~Chen and S.~Wang, ``Regularized boost for semi-supervised learning,'' in
  \emph{Advances in {N}eural {I}nformation {P}rocessing {S}ystems 20}, J.~C.
  Platt, D.~Koller, Y.~Singer, and S.~Roweis, Eds.\hskip 1em plus 0.5em minus
  0.4em\relax Cambridge, MA: MIT Press, 2008, pp. 281--288.

\bibitem{AGA02}
F.~d'Alch{\'e} Buc, Y.~Grandvalet, and C.~Ambroise, ``Semi-supervised
  marginboost,'' in \emph{Advances in {N}eural {I}nformation {P}rocessing
  {S}ystems 14}, T.~G. Dietterich, S.~Becker, and Z.~Ghahramani, Eds.\hskip 1em
  plus 0.5em minus 0.4em\relax Cambridge, MA: MIT Press, 2002, pp. 553--560.

\bibitem{Dietterich2000}
T.~G. Dietterich, ``Ensemble methods in machine learning,'' in
  \emph{Proceedings of the 1st International Workshop on Multiple Classifier
  Systems}, Cagliari, Italy, 2000, pp. 1--15.

\bibitem{FS95}
Y.~Freund and R.~E. Schapire, ``A decision-theoretic generalization of on-line
  learning and an application to boosting,'' in \emph{Lecture {N}otes in
  {C}omputer {S}cience 904}, P.~M.~B. Vit{\'a}nyi, Ed.\hskip 1em plus 0.5em
  minus 0.4em\relax Berlin: Springer, 1995, pp. 23--37.

\bibitem{HB99}
S.~Hettich and S.~D. Bay, ``The {UCI} {KDD} archive [http://kdd.ics.uci.edu],''
  Department of Information and Computer Science, University of California,
  Irvine, CA, Tech. Rep., 1998.

\bibitem{KV95}
A.~Krogh and J.~Vedelsby, ``Neural network ensembles, cross validation, and
  active learning,'' in \emph{Advances in {N}eural {I}nformation {P}rocessing
  {S}ystems 7}, G.~Tesauro, D.~S. Touretzky, and T.~K. Leen, Eds.\hskip 1em
  plus 0.5em minus 0.4em\relax Cambridge, MA: MIT Press, 1995, pp. 231--238.

\bibitem{KW03}
L.~I. Kuncheva and C.~J. Whitaker, ``Measures of diversity in classifier
  ensembles and their relationship with the ensemble accuracy,'' \emph{Machine
  Learning}, vol.~51, no.~2, pp. 181--207, 2003.

\bibitem{Li:Zhou2005}
M.~Li and Z.-H. Zhou, ``{SETRED}: Self-training with editing,'' in
  \emph{Proceedings of the 9th Pacific-Asia Conference on Knowledge Discovery
  and Data mining}, Hanoi, Vietnam, 2005, pp. 611--621.

\bibitem{LZ07}
------, ``Improve computer-aided diagnosis with machine learning techniques
  using undiagnosed samples,'' \emph{IEEE Transactions on Systems, Man and
  Cybernetics - Part A: Systems and Humans}, vol.~37, no.~6, pp. 1088--1098,
  2007.

\bibitem{LJ04}
X.~Lu and A.~K. Jain, ``Ethnicity identification from face images,'' in
  \emph{Proceedings of SPIE International Symposium on Defense and Security},
  Kissimmee, FL, 2004, pp. 114--123.

\bibitem{MJJL09}
P.~K. Mallapragada, R.~Jin, A.~K. Jain, and Y.~Liu, ``Semiboost: Boosting for
  semi-supervised learning,'' \emph{IEEE Transactions on Pattern Analysis and
  Machine Intelligence}, vol.~31, no.~11, pp. 2000--2014, 2009.

\bibitem{MBBF00}
L.~Mason, P.~Bartlett, J.~Baxter, and M.~Frean, ``Functional gradient
  techniques for combining hypotheses,'' in \emph{Advances in {L}arge {M}argin
  {C}lassifiers}, A.~Smola, P.~Bartlett, B.~Sch{\"o}lkopf, and D.~Schuurmans,
  Eds.\hskip 1em plus 0.5em minus 0.4em\relax Cambridge, MA: MIT Press, 2000,
  pp. 221--246.

\bibitem{VJJ08}
H.~Valizadegan, R.~Jin, and A.~K. Jain, ``Semi-supervised boosting for
  multi-class classification,'' in \emph{Proceedings of the 19th European
  Conference on Machine Learning}, Antwerp, Belgium, 2008, pp. 522--537.

\bibitem{WOL92}
D.~H. Wolpert, ``Stacked generalization,'' \emph{Neural Networks}, vol.~5,
  no.~2, pp. 241--259, 1992.

\bibitem{Zhou2009a}
Z.-H. Zhou, ``When semi-supervised learning meets ensemble learning,'' in
  \emph{Proceedings of the 8th International Workshop on Multiple Classifier
  Systems}, Reykjavik, Iceland, 2009, pp. 529--538.

\bibitem{ZL05}
Z.-H. Zhou and M.~Li, ``Tri-training: Exploiting unlabeled data using three
  classifiers,'' \emph{IEEE Transactions on Knowledge and Data Engineering},
  vol.~17, no.~11, pp. 1529--1541, 2005.

\bibitem{ZL09}
------, ``Semi-supervised learning by disagreement,'' \emph{Knowledge and
  Information Systems}, vol.~24, no.~3, pp. 415--439, 2010.

\end{thebibliography}

\end{document}